\documentclass[journal]{IEEEtran}

\usepackage{amsmath,amsfonts}
\usepackage{algorithmic}
\usepackage{algorithm}
\usepackage{array}
\usepackage{textcomp}
\usepackage{stfloats}
\usepackage{url}
\usepackage{verbatim}
\usepackage{graphicx}
\usepackage{cite}
\usepackage{amssymb}
\usepackage{bm}
\usepackage{xspace}
\usepackage{pgfplots}
\usepackage{tabularx}
\usepackage{multirow}
\usepackage{multicol}
\usepackage{threeparttable}
\usepackage{placeins}
\usepackage{balance}

\makeatletter
\let\NAT@parse\undefined
\makeatother
\usepackage[hidelinks]{hyperref}

\newcommand{\reffig}[1]{Fig.~\ref{#1}}
\newcommand{\refeq}[1]{Eq.~\ref{#1}}
\newcommand{\reftab}[1]{Table~\ref{#1}}
\newcommand{\refsec}[1]{Section~\ref{#1}}

\makeatletter
\def\hlinewd#1{%
	\noalign{\ifnum0=`}\fi\hrule \@height #1 \futurelet
	\reserved@a\@xhline}
\makeatother

\usepackage{eso-pic}

\AtBeginDocument{\AddToShipoutPictureFG*{\AtTextUpperLeft{\put(0,\LenToUnit{10pt}){\parbox{\textwidth}{\centering\sffamily IEEE Transactions on Automation Science and Engineering (T-ASE), 2025.
}}}}}

\begin{document}

\title{Dexterous Pre-grasp Manipulation for Human-like Functional Categorical Grasping: Deep Reinforcement Learning and Grasp Representations}

\author{Dmytro Pavlichenko and Sven Behnke
	\thanks{Both authors are with the Autonomous Intelligent Systems (AIS)
		Group, Computer Science Institute VI -- Intelligent Systems and Robotics -- and the Center for Robotics and the Lamarr Institute for Machine Learning and Artificial Intelligence, University of Bonn, Germany. {\tt pavlichenko@ais.uni-bonn.de}}
	\thanks{This work was funded by the German Ministry of Education and
		Research (BMBF), grant no. 01IS21080, project "Learn2Grasp: Learning
		Human-like Interactive Grasping based on Visual and Haptic Feedback".}
}

\maketitle

\begin{abstract}
Many objects, such as tools and household items, can be used only if grasped in a very specific way---grasped functionally. Often, a direct functional grasp is not possible, though. We propose a method for learning a dexterous pre-grasp manipulation policy to achieve human-like functional grasps using deep reinforcement learning. We introduce a dense multi-component reward function that enables learning a single policy, capable of dexterous pre-grasp manipulation of novel instances of several known object categories with an anthropomorphic hand. The policy is learned purely by means of reinforcement learning from scratch, without any expert demonstrations. It implicitly learns to reposition and reorient objects of complex shapes to achieve given functional grasps. In addition, we explore two different ways to represent a desired grasp: explicit and more abstract, constraint-based. We show that our method consistently learns to successfully manipulate and achieve desired grasps on previously unseen object instances of known categories using both grasp representations. Training is completed on a single GPU in under three hours.
\end{abstract}

\def\abstractname{Note to Practitioners}
\begin{abstract}
This work was motivated by the increasing popularity of robots equipped with dexterous human-like hands. Operating in environments designed for humans necessitates the ability to use human tools. That requires grasping these tools in specific ways for effective use. We propose a learning-based method to train such behaviors in highly parallelized simulation. We explore two possible ways to represent a target functional grasp: an explicit and a more abstract, constraint-based, each with its own advantages and disadvantages. Our method learns to achieve human-like behaviors in under three hours on a single computer. It successfully manipulates previously unseen object instances with both target grasp representations. Such policies could be useful for robots with human-like hands in a broad range of scenarios: household, factory or search-and-rescue, whenever there is a necessity to grasp objects in a very specific way. The main limitation of this work is that the learned behaviors were not tested in the real world. Thus, closing the sim-to-real gap is a viable direction for future work.
\end{abstract}
\def\abstractname{Abstract}


\section{Introduction}
\label{sec:Introduction}

Grasping is a fundamental skill that manipulation robots need to interact with their environment. Many objects are made for human hands and require a specific grasp to be utilized. For example, a drill requires a power grasp with the index finger on the trigger. We refer to such grasps as \textit{functional grasps}. Often, a functional grasp cannot be achieved directly because the object is in the wrong pose. This can be addressed with \textit{pre-grasp manipulation}: repositioning and reorienting the object until the desired functional grasp is achieved. Robustly performing interactive functional grasping with a dexterous multi-finger hand is challenging, though. Solving this tasks is an important step towards enabling robots to use the tools and functional objects designed for humans.

\begin{figure}[t]
	\vspace*{-1ex}
	\centering
	\includegraphics[width=0.32\linewidth]{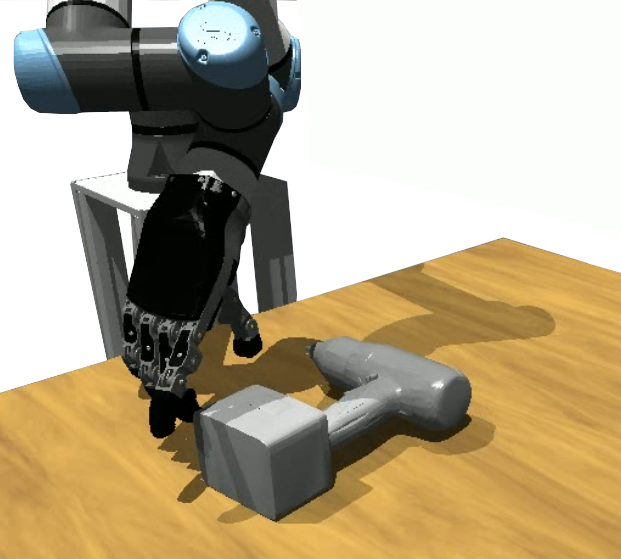}
	\includegraphics[width=0.32\linewidth]{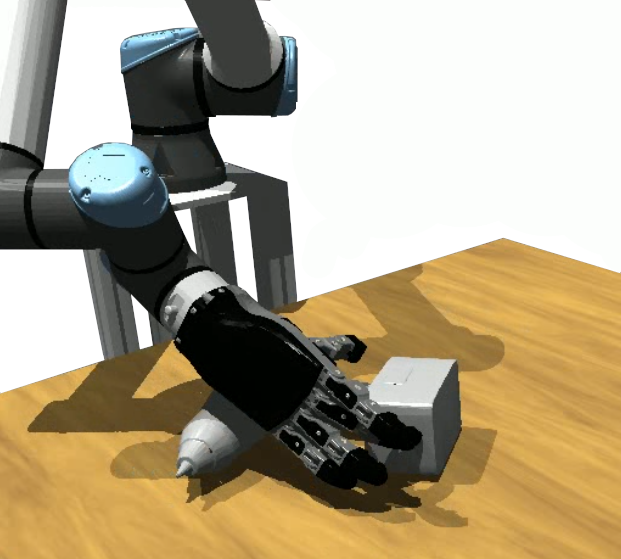}
	\includegraphics[width=0.32\linewidth]{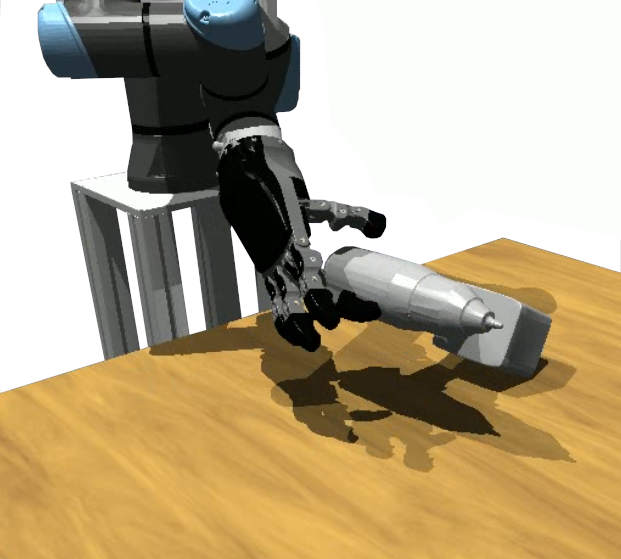}
	\\ \vspace*{1ex}
	\includegraphics[width=0.32\linewidth]{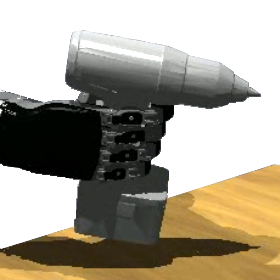}
	\includegraphics[width=0.32\linewidth]{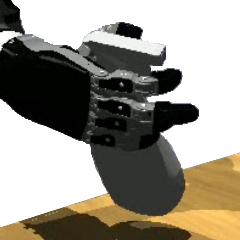}
	\includegraphics[width=0.32\linewidth]{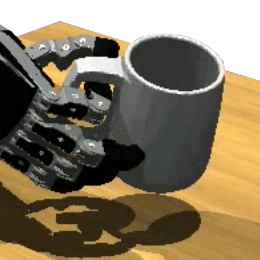}
	\caption{\emph{Top:} Dexterous pre-grasp manipulation that includes reorienting and repositioning a drill. \emph{Bottom:} Provided only with a target index fingertip position and desired object orientation, our policy learned to utilize a human-like hand to achieve intuitive grasps for three object categories.}
	\label{fig:teaser}
	\vspace*{-3ex}
\end{figure}

Inspired by our previous work on functional re-grasping~\cite{Pavlichenko_2019}, we propose a Deep Reinforcement Learning (DRL)-based methodology that replaces several complex classical components with a single data-driven approach. 

DRL has been applied to several challenging robotic domains~\cite{Hwangbo_2019, Rodriguez_2021, Chen_2021, Babuska_2019}. In this work, we use a highly efficient GPU-based simulation~\cite{Makoviychuk_2021} together with DRL to learn a policy for dexterous pre-grasp manipulation. Many approaches focus on learning the policies directly from low-level sensory inputs, such as camera images and point clouds~\cite{Mandikal_2020, Qin_2022}. We observe, however, that most of the data points in these inputs, such as background pixels in an image, are irrelevant to the manipulation policy. Therefore, we assume that perception is performed by an external method, and our approach is provided with high-level semantic information.

The perception task can be addressed by reconstructing the object shape from partial observations~\cite{Rodriguez_2020, Han_2021, Yang_2021, Zhou_2024}, transferring the functional grasp~\cite{Rodriguez_2018, Wu_2023, Zhang_2023, Wei_2024}, and estimating the 6D pose of the object~\cite{Deng_2020, Hu_2020, Hofer_2021, AminiPB:IAS22, Wang_2024}. External perception speeds up the learning process since the policy can be represented by a model with fewer parameters. Additionally, expensive image rendering is avoided. 

By considering multiple object instances within the same category, we further reduce the inputs to the policy, as category-specific features of object geometry and dynamics are learned implicitly. 

Finally, we eliminate the need for expert demonstrations by introducing a dense multi-component reward function that naturally encourages dexterous manipulation.

An achieved desired functional grasp is the final product of the learned policy. The target grasp representation greatly influences the speed of learning and the final result. In this work, we explore two different target grasp representations, having different strengths and weaknesses.

First, the \textit{explicit target grasp representation} strictly defines the desired hand and finger poses relative to an object. It introduces a clear desired result and prevents the policy from being stuck in sub-optimal behaviors. On the other hand, this representation needs an external oracle defining these explicit targets, which may be challenging for novel object categories.

Second, the \textit{constraint-based target grasp representation} is more abstract and is defined with a constraint. In this work, we use index fingertip position and hand orientation relative to an object. The constraint-based representation is more compact and enables the policy to learn a variety of different grasps, satisfying the constraint. Defining this constraint for novel instances is an easier task compared to an explicit grasp representation. However, this comes at the cost of ensuring that the learned grasps are sufficient to securely lift and utilize the objects. We show that the proposed dense multi-component reward function can be effectively applied to both grasp representations with minimal modifications and consistently yields meaningful policies.

To evaluate the proposed method, we learn a single policy in simulation on three conceptually distinct rigid object categories: drills, spray bottles, and mugs. Two proposed target grasp representations are utilized, and the corresponding policies are evaluated. Using dense multi-component reward, the policy learns to perform dexterous pre-grasp manipulation on previously unseen object instances of known categories with a high success rate. \reffig{fig:teaser} shows examples of the learned behaviors. Learning is performed in simulation in less than three hours on a single GPU. This work is an extension of our previous research~\cite{Pavlichenko_2023}. The main contributions of this article are:
\begin{itemize}
	\item a constraint-based target functional grasp representation -- this representation enables exploring different grasp configurations -- and 	%
	\item a multi-component dense reward formulation that quickly yields policies capable of dexterous manipulation.
\end{itemize}
\section{Related Work}
\label{sec:Related_work}

Dexterous pre-grasp manipulation has been an active area of research for decades. Multiple classical model-based approaches have been proposed~\cite{Han_1998, Dogar_2010, Chang_2010, Moll_2018, Hang_2019}. They work for known objects with exact models but require carefully hand-crafted, task-dependent algorithms and suffer from uncertainty inherent in dexterous and highly dynamic manipulation.

In our previous research~\cite{Pavlichenko_2019}, we address functional grasping by means of re-grasping with a dual-arm robot. The manipulation pipeline is implemented with several classical approaches. While humans commonly employ both hands simultaneously, the dexterity of a single human hand significantly exceeds the requirements for functional pre-grasp manipulation. A natural progression toward achieving comparable performance with robotic manipulators involves executing the same task using a single hand. However, achieving such highly dynamic manipulation with classical approaches is very challenging.

A promising solution to such problems is to leverage data-driven methods. In particular, DRL and Imitation Learning (IL) using Artificial Neural Networks (ANN) to represent policies for dexterous manipulation have gained much popularity in recent years~\cite{Zhu_2019, openai_2020, Levine_2016}. By learning purely from observed experiences and/or provided demonstrations, these methods yield interactive policies capable of dexterous multi-finger manipulation.

Zhou et al.~\cite{Zhou_2022} address pre-grasp manipulation of objects in ungraspable configurations through extrinsic dexterity. Their method uses model-free RL to learn to push objects against a wall to achieve a graspable pose. The method uses a minimalistic object representation, similar to our approach. However, it has difficulties generalizing to objects with complex non-convex shapes. In our approach, this issue is resolved by learned implicit category-specific geometry knowledge. Similarly, Sun et al.~\cite{Sun_2020} use model-free RL to obtain a policy for a dual-arm robot that pushes an object next to a wall and turns it to grasp it with the other hand. Both works use parallel grippers, which make the manipulation less dexterous.

Yuzhe et al.~\cite{Qin_2022} train a dexterous manipulation policy for an Allegro hand to grasp novel objects of a known category. Their approach uses point clouds as input to provide information about object geometry. The difference to our work is that we specifically address functional grasping, while in their work the grasps are arbitrary. Mandikal et al.~\cite{Mandikal_2020} propose to learn a policy with object-centric affordances to dexterously grasp objects. Notably, the policy is learned with a prior derived from observing manipulation videos, which requires tedious annotation of human grasp regions in the observed images.

A wide range of works is based on real-world expert demonstrations~\cite{Aravind_2018, Radosavovic_2021, Mandikal_2022}. These approaches have to deal with the challenges of mapping human motion to the kinematics of the robot arm and hand. Hence, the direct applicability to different robotic setups is not straightforward. Chen et al.~\cite{Chen_2022} address this issue by bootstrapping a small dataset of human demonstrations with a larger dataset including novel objects and grasps. The objects are deformed and dynamically consistent grasps are generated. The policy is then trained in a supervised manner in simulation, followed by a direct transfer to the real world. This method struggles with objects of complex shapes. Palleschi et al.~\cite{Palleschi_2023} utilize human demonstrations to teach the policy to grasp a wide range of objects. The approach is evaluated on two grippers: soft and rigid. In contrast, in our work, we avoid using explicit demonstrations and instead rely on a general and dense reward function to guide the policy towards dexterous manipulation.

A completely different approach was proposed by Dasari et al.~\cite{Dasari_2022}: a combination of exemplar object trajectories with predefined pre-grasp configurations as training data. The pre-grasp-based approaches were also introduced in~\cite{Kappler_2010, Baek_2021, Kappler_2012}. Our method shares the high-level idea with these works. The policy learns to perform a wide variety of tasks in simulation without any task-specific engineering. The key difference to our work is that learned behaviors directly depend on supplied exemplar trajectories. In addition, the manipulation is performed by a freely floating hand, which relaxes multiple constraints introduced by the kinematics of the robotic arm in combination with object poses on the edge of the workspace.

Wu et al.~\cite{Wu_2024unidexfpm} utilize a teacher-student approach with policy distilling to obtain a policy that is capable of repositioning and reorienting objects of a vast range of categories to achieve required functional grasp object poses without grasping the object. In contrast, our approach focuses on achieving required functional grasps while simultaneously learning to perform the pre-grasp manipulation.

Agarwal et al.~\cite{Agarwal_2023dexterous} propose an object-hand manipulation representation for dexterous robotic hands, followed by a functional grasp synthesis framework, and evaluate the approach in the real world. The main disadvantage of this method is the necessity to compose a complex and large dataset. Zhu et al.~\cite{Zhu_2023} propose to use Eigengrasps to reduce the search space of RL using a small dataset collected from human expert demonstrations. The target functional grasps are predicted with an affordance model. The approach is successfully transferred and evaluated in the real world. In contrast to these approaches, we avoid any expert demonstrations and focus on obtaining dexterous manipulation policies solely through a dense multi-component reward function that is formulated without using specific hand, arm, or object details.
\section{Background}
\label{sec:Background}

The objective of this work is to learn a policy $\pi$ that achieves the desired behavior: pre-grasp manipulation of novel object instances with the aim of reaching a functional grasp. The policy $\pi_{\theta}$ is represented by a deep neural network and is parameterized by weights $\theta$, learned with DRL. The problem is modeled as a Markov Decision Process (MDP): $\{S,A,P,r\}$ with state space $S \in \mathbb{R}^{n}$, action space $A \in \mathbb{R}^{m}$, state transition function $P \colon S \times A \mapsto S$, and reward function $r \colon S \times A \mapsto \mathbb{R}$. Since the problem has continuous state and action space, the policy $\pi_{\theta}(\bm{a}|\bm{s})$ represents an action probability distribution when observing a state $\bm{s}(t)$ at timestep $t$. 

The objective of DRL is to maximize the \textit{expected return}:
\begin{equation}
	J(\pi_{\theta}) = \sum_{t=0}^{T} \mathbb{E} [ \gamma^t r(\bm{s}(t), \bm{a}(t))],
\end{equation}
where $\gamma \in [0,1]$ is a discounting factor.

The policy is provided with a target functional pre-grasp to reach, defined as a 6D hand pose in an object frame plus hand joint positions in case of the explicit target grasp representation. In case of the constraint-based target grasp representation, the policy is provided with a target position for the index fingertip and end-effector orientation. 

We assume that there is a single object in front of the robot hand on the table.

\section{Explicit Target Grasp Representation}
\label{sec:Method}

A target functional grasp is explicitly represented by a 6D pose of the end-effector relative to an object and the joint positions of the fingers. This representation has the advantage of providing the policy with a concrete goal, which typically enhances learning speed and convergence stability. Disadvantages of such representations are twofold. First, very specific grasps that lead to a desired outcome have to be produced, which is not straightforward. Second, an explicit target grasp disallows the policy to explore other grasp configurations, satisfying the constraint that defines a grasp as functional. 

In this section, we present the methodology for learning pre-grasp manipulation with explicit grasp representation.

\subsection{State Space}
\label{sec:State_1}

\begin{figure*}[t]
	\centering
	\includegraphics[width=\linewidth]{figures/diagram.pdf}
	\caption{Composition of the state representation and the reward function. The state consists of information about the hand, the object, and the target functional grasp. The reward function consists of a term encouraging reaching the target grasp, a term encouraging pre-grasp manipulation, and a low manipulability score penalty. "O" denotes object frame of reference.}
	\label{fig:diagram}
	\vspace*{-3ex}
\end{figure*}

The \textit{state vector} $\bm{s}(t)$ consists of three distinctive parts: information about the hand $\bm{h}$, about the object $\bm{o}$, and about the target functional grasp $\bm{g}$:
\begin{equation}
	\bm{s} = [\bm{h}, \bm{o}, \bm{g}].
\end{equation}
The left part of \reffig{fig:diagram} illustrates the state representation.

Information about the current \textit{state of the hand} is a column vector:
\begin{equation}
	\bm{h} = [\bm{h}_\textrm{p}, \bm{h}_\textrm{r}, \bm{h}_\textrm{j}, \bm{h}^O_\textrm{p}, \bm{h}^O_\textrm{r}],
\end{equation}
where  $\bm{h}_\textrm{p} = [h_{\textrm{p}_x}, h_{\textrm{p}_y}, h_{\textrm{p}_z}]$ is a 3D hand position vector, $\bm{h}_\textrm{r}$ is a 4-element hand rotation vector represented by a quaternion, and $\bm{h}_\textrm{j}$ is a 5-element hand joint position vector; $\bm{h}^O_\textrm{p}$ and $\bm{h}^O_\textrm{r}$ are hand position and rotation in the object frame of reference $O$. Thus, information about the hand $\bm{h}$ is a 19-element vector. Throughout this article, we frequently use the following subscripts: $x_\textrm{p}$ denoting 3D position, $x_\textrm{r}$ denoting rotation expressed as a quaternion, and $x_\textrm{j}$ denoting joint positions.

There is a degree of redundancy in the hand pose, since it is included in two reference frames: in global and object frames. Having a global hand pose is useful for the method to better understand how close it is to the workspace boundaries. At the same time, hand pose relative to the object is more representative for the spatial understanding of manipulation. While one can be inferred from the other, that would prolong the learning process for the sake of learning something that is already known, which we deem to be infeasible. 

As opposed to action space~(\refsec{sec:Action_1}), in the state space we define rotations with quaternions. Thanks to their property of being unique, they avoid making the policy learn to navigate the Euler angles space at a cost of one extra dimension.

\textit{Information about the object} is a column vector:
\begin{equation}
	\bm{o} = [\bm{o}_\textrm{p}, \bm{o}_\textrm{r}, \bm{o}_\textrm{bb}, \bm{o}_\textrm{s}, \bm{o}_\textrm{c}],
\end{equation}
where $\bm{o}_\textrm{p}$ is 3D object position, $\bm{o}_\textrm{r}$ is a 4-element object rotation vector represented by a quaternion, $\bm{o}_\textrm{bb}$ is a 6-element vector representing the object bounding box by two 3D positions of diagonally opposing bounding box corners, $\bm{o}_\textrm{s}$ is a 10-element vector of signed distances between fingertips and middles of the fingers to the object surface, and $\bm{o}_\textrm{c}$ is a C-element one-hot vector representing the object category. 

Distances from fingers to object surface are efficiently calculated from a pre-computed object Signed Distance Field (SDF)~\cite{Mosbach_2022}. Thus, information about the object is a (23+C)-element vector. Such representation is compact, leveraging the known category to implicitly learn typical features of object instances.

The \textit{desired functional grasp} is provided as a column vector:
\begin{equation}
	\bm{g} = [\bm{g}^O_\textrm{p}, \bm{g}^O_\textrm{r}, \bm{g}_\textrm{j}],
\end{equation}
where $\bm{g^O}_\textrm{p}$ is 3D hand position in the object frame of reference, $\bm{g}^O_\textrm{r}$ is a 4-element hand rotation vector represented by a quaternion, and $\bm{g}_\textrm{j}$ is a 5-element hand joint position vector. 

The target functional grasp is represented by a 12-element vector. In practice, functional grasps can be provided by~\cite{Rodriguez_2018, Zhu_2021}. Grasp $\bm{g}$ is kept fixed during a trial.

In this work, we consider $C=3$ categories. Thus, the state is a 57-element vector. It resembles a high-level semantic representation of the scene. This compact state can be computed fast on a GPU and thus facilitates quick learning. Moreover, compared to DRL models that learn directly from raw visual inputs, smaller models with fewer parameters can be used. 

\subsection{Action Space}
\label{sec:Action_1}

The policy produces actions $\bm{a}(t)$ with a frequency of 30\,Hz. An action represents a relative displacement in 3D hand position, hand rotation, and hand joint positions. With this action definition, hand joint targets are straightforward to obtain. The arm joint targets are calculated via Inverse Kinematics (IK). Finally, the joints are controlled with PD controllers.

In this work, we apply the proposed method to a 6\,DoF UR5e robotic arm with a 11\,DoF Schunk SIH hand. The joints of the hand are coupled, leaving five controllable DoF. Thus, in this work an action is an 11-element vector: three elements define a displacement of hand position, three elements define a displacement of hand rotation as Euler angles, and five elements define a displacement of hand joint positions. We further assume a five-fingered hand with five controllable DoF. However, it is straightforward to apply our approach to a hand with an arbitrary number of DoF. We use Euler angles representation for the rotation-related part of the action since it is straightforward to obtain the next target with small iterative increments while using a minimal number of variables.

\subsection{Reward Function}
\label{sec:Reward_Function}

The right part of \reffig{fig:diagram} illustrates the composition of the reward function $r(t)$ that is defined as:
\begin{equation}
	\label{eq:reward}
	r(t) = r_{\textrm{grasp}}(t) + r_{\textrm{man}}(t) + r_{\textrm{MP}}(t) + r_\textrm{T}(t),
\end{equation}
where $r_{\textrm{grasp}}$ encourages movement towards the target grasp $\bm{g}$, $r_{\textrm{man}}$ encourages pre-grasp manipulation of an object, $r_{\textrm{MP}}$ penalizes being in configurations with low manipulability, and $r_\textrm{T}$ rewards reaching the target functional grasp $\bm{g}$. Each reward component is defined to be in $[-1, 1]$ and described in detail below. For brevity, we omit specifying a dependency on time $t$, unless necessary.

First, we define the distance function $\phi$ between two quaternions $\bm{q}$ and $\bm{q}'$ as the rotation between them:
\begin{equation}
	\phi(\bm{q}, \bm{q}') = 2 \arccos((\bm{q} \cdot \bm{q}'^{-1})_4).
	\label{eq:angular_distance}
\end{equation}
%

The \textit{grasp reward} $r_{\textrm{grasp}}$ is defined as:
\begin{equation}
	r_{\textrm{grasp}} = r_{\bm{h}_\textrm{p}} + r_{\bm{h}_\textrm{r}} + \lambda r_{\bm{h}_\textrm{j}},
\end{equation}
where $r_{\bm{h}_\textrm{p}}$ encourages moving the hand position towards the target 3D grasp position, $r_{\bm{h}_\textrm{r}}$ encourages moving the hand rotation towards the target grasp rotation, and $r_{\bm{h}_\textrm{j}}$ encourages moving hand joint positions towards target grasp joint positions. $\lambda \in [0,1]$ is the grasp joint reward importance factor. 

Overall, the $r_{\textrm{grasp}}$ reward encourages aligning hand pose and joint positions with the target grasp pose and joint positions.

The \textit{hand position reward} $r_{\bm{h}_\textrm{p}}$ is defined as:
\begin{equation}
	r_{\bm{h}_\textrm{p}}(t) = \frac{\Delta \bm{h}_\textrm{p}(t-1) - \Delta \bm{h}_\textrm{p}(t)}{\Delta \bm{h}_\textrm{p}^\textrm{max}},\;\;
	\label{eq:reward_hand}
	\Delta \bm{h}_\textrm{p} = || \bm{h}^O_\textrm{p} - \bm{g}^O_\textrm{p} ||,
\end{equation}
where $\Delta \bm{h}_\textrm{p}$ is the Euclidean distance from the hand position $\bm{h}^O_\textrm{p}$ to the target grasp hand position $\bm{g}^O_\textrm{p}$. $\Delta \bm{h}_\textrm{p}^\textrm{max}$ is a maximal hand position change during the step duration $\Delta t$: $\Delta \bm{h}_\textrm{p}^\textrm{max} = v_{\bm{h}_\textrm{p}}^\textrm{max}\Delta t$ with $v_{\bm{h}_\textrm{p}}^\textrm{max}$ being the maximal linear velocity of the hand. In the case of the UR5e manipulator used in this work, $v_{\bm{h}_\textrm{p}}^\textrm{max} = 1$\,m/s and $\Delta t = 0.0333$\,s.

The \textit{hand rotation reward} is defined as:
\begin{equation}
	r_{\bm{h}_\textrm{r}}(t) = \frac{\Delta \bm{h}_\textrm{r}(t-1) - \Delta \bm{h}_\textrm{r}(t)}{\Delta \bm{h}_\textrm{r}^\textrm{max}},\;\;
	\Delta \bm{h}_\textrm{r} = \phi(\bm{h}^O_\textrm{r}, \bm{g}^O_\textrm{r}),
\end{equation}
where $\Delta \bm{h}_\textrm{r}$ is a distance from the hand rotation $\bm{h}^O_\textrm{r}$ to the target grasp hand rotation $\bm{g}^O_\textrm{r}$, calculated according to \refeq{eq:angular_distance}. $\Delta \bm{h}_\textrm{r}^\textrm{max}$ is a maximal hand rotation change during time $\Delta t$. It is defined analogously to $\Delta \bm{h}_\textrm{p}^\textrm{max}$. We use $v_{\bm{h}_\textrm{r}}^\textrm{max} = \pi$\,rad/s.

Finally, the \textit{hand joint reward} is defined as:
\begin{equation}
	\!\!r_{\bm{h}_\textrm{j}}(t)\!=\!\frac{\Delta \bm{h}_\textrm{j}(t\!-\!1) - \Delta \bm{h}_\textrm{j}(t)}{\Delta \bm{h}_\textrm{j}^\textrm{max}},
	\Delta \bm{h}_\textrm{j}\!=\!\frac{1}{N} \sum_{i=0}^{N}|h_{\textrm{j}_i}\!-\!g_{\textrm{j}_i}|,
\end{equation}
where $N$ is the number of controllable hand joints, $\Delta \bm{h}_\textrm{j}$ is an average per-joint distance to the target grasp joint positions, and $\Delta \bm{h}_\textrm{j}^\textrm{max}$ is a maximal joint position displacement during time $\Delta t$. It is defined similarly to the maximal position and rotation displacements through maximal joint velocity. We use $v_{\bm{h}_\textrm{j}}^\textrm{max} = \pi$\,rad/s.

The \textit{hand joint importance factor} $\lambda$ is defined as:
\begin{equation}
	\!\!\lambda \!=\! \Big( 1 - \frac{\min(h_\textrm{p}^{\textrm{prox}}, \Delta \bm{h}_\textrm{p})}{h_\textrm{p}^{\textrm{prox}}} \Big) \Big( 1 - \frac{\min(h_\textrm{r}^{\textrm{prox}}, \Delta \bm{h}_\textrm{r})} {h_\textrm{r}^{\textrm{prox}}} \Big),\!
\end{equation}
where $h_\textrm{p}^{\textrm{prox}}$ is a predefined constant, representing a proximity distance between the hand position and the target grasp position, from which the hand joint position reward becomes active. We set it to the length of the hand. Similarly, $h_\textrm{r}^{\textrm{prox}}$ is a rotation proximity distance; we use $h_\textrm{r}^{\textrm{prox}} = 1$\,rad. 

Overall, $\lambda$ leads to ignoring the hand joint reward when the hand is far from the target grasp pose. Instead, the use of fingers for manipulation is promoted.

\begin{figure*}[t]
	\centering
	\includegraphics[width=0.9\linewidth]{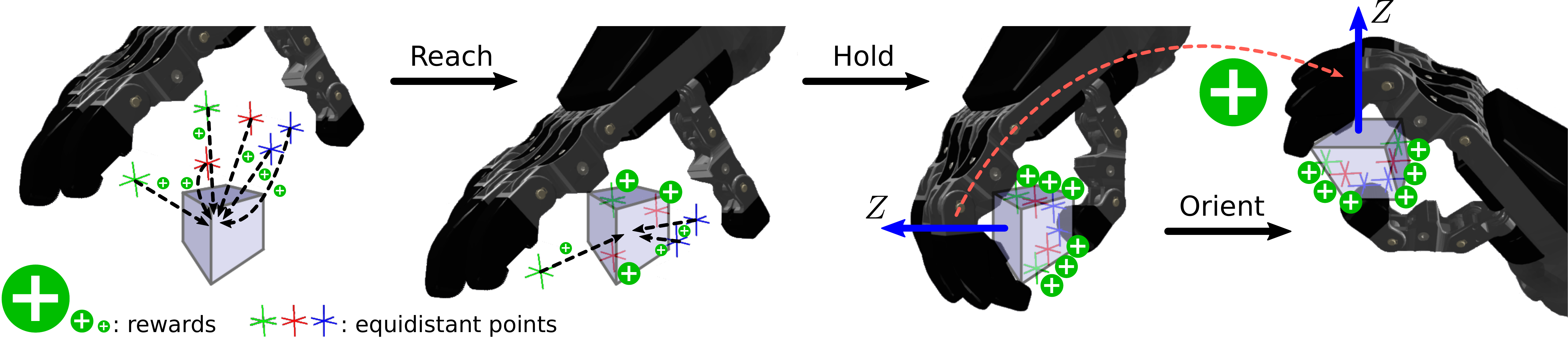}
	\caption{Manipulation reward $r_{\textrm{man}}$ composed of three components: reach, hold, and orient, representing a sequence of interconnected tasks. Equidistant points between the thumb tip and middle fingertip \& center, used to query distances to the object, are red, blue, and green crosses. Rewards: plus signs, size is proportional to reward value. First, the motion of the equidistant points to the object surface is rewarded by the reach reward. Second, equidistant points that are inside the object yield a bigger hold reward. Finally, orienting the object towards the nominal orientation yields an even bigger reward. Note that closing the hand brings the equidistant points closer together, often pushing them inside the object. This design implicitly rewards grasping behaviors without using expensive contact information or explicitly rewarding specific movement primitives.}
	\label{fig:manipulation_reward_diagram}
\end{figure*}

The \textit{manipulation reward} $r_{\textrm{man}}$ is defined as:
\begin{equation}
	r_{\textrm{man}} = r_\textrm{reach} + r_\textrm{hold} + r_\textrm{orient},
\end{equation}
where $r_\textrm{reach}$ encourages moving the hand towards the object, $r_\textrm{hold}$ encourages holding the object in the hand, and $r_\textrm{orient}$ encourages orienting the object towards a nominal rotation, where the target grasp is more likely to be reachable. Thus, the manipulation reward encourages a canonical \emph{reach} $\rightarrow$ \emph{hold} $\rightarrow$ \emph{orient} behavior for pre-grasp object manipulation. The manipulation reward is illustrated in \reffig{fig:manipulation_reward_diagram}. All its terms are strictly positive.

The \textit{hand reach reward} is defined as:
\begin{equation}
	r_\textrm{reach}(t) = \frac{\sum_{k=1}^{K}\big(d(\bm{H}_{\textrm{p}_k}(t-1)) - d(\bm{H}_{\textrm{p}_k}(t))\big)}{\Delta \bm{h}_\textrm{p}^\textrm{max}},
\end{equation}
where $d$ is a function, taking a set of 3D points and returning signed distances from the points to the object surface, utilizing the precomputed object SDF. $\Delta \bm{h}_\textrm{p}^\textrm{max}$ is a maximal position displacement, defined in \refeq{eq:reward_hand}. $\bm{H}_\textrm{p}$ is a set of $K$ 3D points between the thumb and the other fingers, described in detail below. In the context of this reward, these points guide the hand towards a position where the object is between the thumb and the other fingers, which is advantageous for manipulation.

The \textit{object hold reward} is defined as:
\begin{equation}
	r_\textrm{hold} = \frac{1}{K} \sum_{k=1}^{K} \frac{d(\bm{H}_{\textrm{p}_k}) - \rho}{d_k^\textrm{max}},
\end{equation}
where $\rho$ is a predefined constant radius of spheres with points $\bm{H}_\textrm{p}$ as centers and $d_k^\textrm{max}$ is a per-point maximum possible distance from the point to the closest finger surface. The set of hold-detect points $\bm{H}_\textrm{p}$ is positioned between the tip of the thumb and the tip or middle of the other fingers. Thus, points between fingertips represent positions where objects can be pinch-grasped and points between the thumb-tip and finger-middles represent positions where objects are grasped more securely.

Each direction thumb-tip$\,\rightarrow\,$fingertip and thumb-tip$\,\rightarrow\,$finger-middle has three equidistant points. This ensures a positive response when an object is positioned between the thumb and other fingers imperfectly. When the hand closes, the equidistant points come closer to each other. This promotes closing the hand around an object. Note that the maximum $r_\textrm{hold}$ is achieved when fingers evenly embrace the object, which naturally resembles a good grasp.

For simplicity, we use only the points between the thumb and the middle-finger in this work, yielding six points, as shown in \reffig{fig:manipulation_reward_diagram}. The intuition behind this design choice is that if an object is contained between the middle finger and the thumb, it is also contained between the index and ring fingers, as defined by the hand topology. While it is straightforward to utilize several finger pairs at the same time, we observed in practice that using only the middle-thumb line was sufficient to learn grasping behaviors. Having all other fingertips close to the object is encouraged by the reach reward term $r_\textrm{reach}$.

The \textit{object orient reward} is defined as:
\begin{equation}
	\!\!\!r_\textrm{orient}(t)\!= \frac{\Delta \bm{o}_\textrm{r}(t\!-\!1)-\Delta \bm{o}_\textrm{r}(t)}{\pi},
	\Delta \bm{o}_\textrm{r}\!=\!\phi(\bm{o}_\textrm{r}, \bm{o}_\textrm{r}^\textrm{nominal}),
\end{equation}
where $\Delta \bm{o}_\textrm{r}$ is the distance from the object rotation to the nominal object rotation $\bm{o}_\textrm{r}^\textrm{nominal}$. A nominal rotation resembles a natural object orientation as intended for functional use: the object z-axis points upwards, and the object x-axis (the direction of the tool tip) points away from the hand. Although many other object orientations would be feasible to perform a functional grasp, we find that such a definition is generic and unbiased. In practice, $r_\textrm{orient}$ provides good guidance on how to reorient an object when it is in a state where a direct functional grasp is not possible. 

Together, the reach, hold, and orient rewards represent a sequence of interconnected tasks that help to steer the policy towards dexterous manipulation behaviors.

The \textit{manipulability penalty reward} is defined as:
\begin{equation}
	r_{\textrm{MP}} = 1 - 2 \Big/ \Big( {1 + \bigl( \tfrac{\min(|\bm{J}|, |\bm{J}|_\textrm{max})}{|\bm{J}|_\textrm{max}} \bigr) ^3} \Big) ,
\end{equation}
where $|\bm{J}|$ is a determinant of the end-effector Jacobian $\bm{J}$ and $|\bm{J}|_\textrm{max}$ is a maximum determinant value that is penalized. We define $|\bm{J}|_\textrm{max}$ to be 15\% of maximal observed $|\bm{J}|$ for a specific arm. This reward penalizes coming close to singularities and leads to learning more intuitive motions.

Finally, the \textit{target grasp reward} is defined as:
\begin{equation}
	\label{eq:target_1}
	r_\textrm{T} =
	\begin{cases}
		1 & \text{if $\Delta \bm{h}_\textrm{p} < T_\textrm{p} \wedge 
			\Delta \bm{h}_\textrm{r} < T_\textrm{r} \wedge 
			\Delta \bm{h}_\textrm{j} < T_\textrm{j}$} \\
		0 & \text{otherwise,}
	\end{cases}
\end{equation}
where $T_\textrm{p}, T_\textrm{r}, T_\textrm{j}$ are the distance thresholds for hand position, rotation, and hand joint positions to the target grasp that define the accuracy with which the target grasp is achieved. We use $T_\textrm{p}=1$\,cm, $T_\textrm{r}=0.15$\,rad, and $T_\textrm{j}=0.1$\,rad. The episode ends when reaching the target grasp. 

Wide use of differential distances in our reward, instead of directly using the velocities, naturally avoids learning overshooting behaviors. Note that all reward terms are defined in a generic way and can be easily configured for an arbitrary robotic arm and hand. All reward components are defined to be in the interval $[-1, 1]$ or $[0, 1]$. This allows applying relative scaling easily. For best performance, we scale the rewards according to their position in the sequence of interconnected sub-tasks, as shown in \reffig{fig:manipulation_reward_diagram}: $r_\textrm{T} \gg r_\textrm{orient} \gg r_\textrm{hold} \gg r_\textrm{reach}$. We leave the other rewards unscaled. This reduces the probability that the policy gets stuck in the local minima, created by accumulating rewards for actions that are easier to achieve compared to the following more complex sub-tasks.

To summarize, the multi-component reward function can be split into three terms:
\begin{enumerate}
	\item the manipulability penalty reward $r_{\textrm{MP}}$ penalizes being close to singularities and thus helps to avoid unintuitive behavior,
	\item the grasp reward $r_{\textrm{grasp}}$ encourages reaching the given functional grasp, and
	\item the manipulation reward $r_{\textrm{man}}$ encourages reaching, holding, and reorienting the object.
\end{enumerate}
Each component is a continuous dense reward. The components combine to effectively guide the policy towards learning a robust, dexterous pre-grasp manipulation. Finally, a sparse component $r_\textrm{T}$ rewards reaching the target grasp.

\subsection{Curriculum}
\label{sec:Curriculum}

In this work, we avoid having any explicit expert demonstrations and focus on learning robust and natural policies for object pre-grasp manipulation through pure DRL with dense reward shaping. To facilitate faster and more stable learning, we propose a simple two-stage curriculum.

In the first stage, we place the objects in poses where target functional grasps can be reached directly. 
Objects are positioned on the table in their nominal poses 5\,cm away from the inner side of the hand. The arm is set to a neutral configuration with a high manipulability score. We disable the $r_{\textrm{man}}$ reward term during the first stage so that the policy can converge faster. 

The second stage has full difficulty, taking advantage of the warm-start provided by the first stage. The curriculum is agnostic to object-specific details, keeping the approach general while achieving faster policy convergence.
\section{Constraint-based Target Grasp Representation}
\label{sec:Method_2}

A target functional grasp can be represented using a constraint that defines the grasp as functional. For instance, with a drill, such a constraint might be positioning the index fingertip on the trigger to make the activation of the drill possible. 

\begin{figure}[t]
	\centering
	\includegraphics[width=0.48\linewidth]{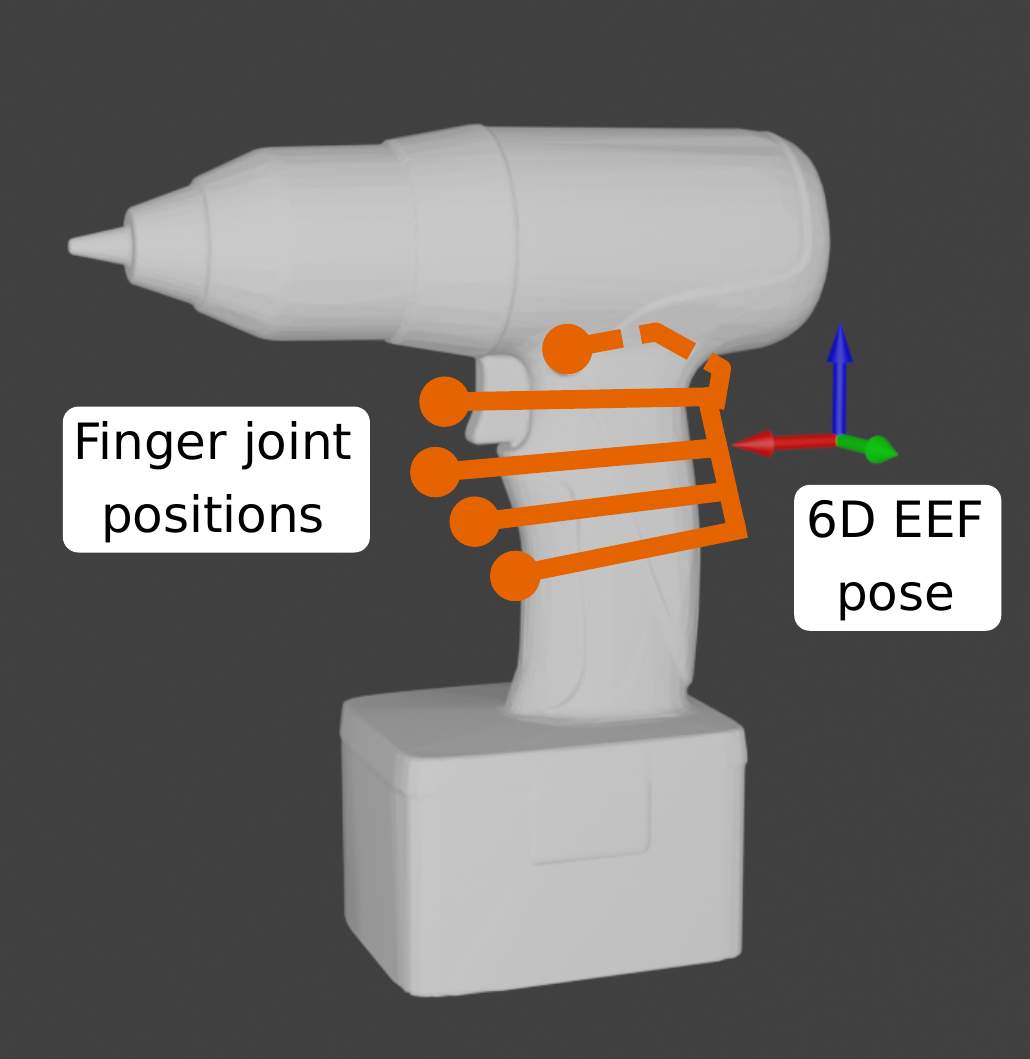}
	\includegraphics[width=0.48\linewidth]{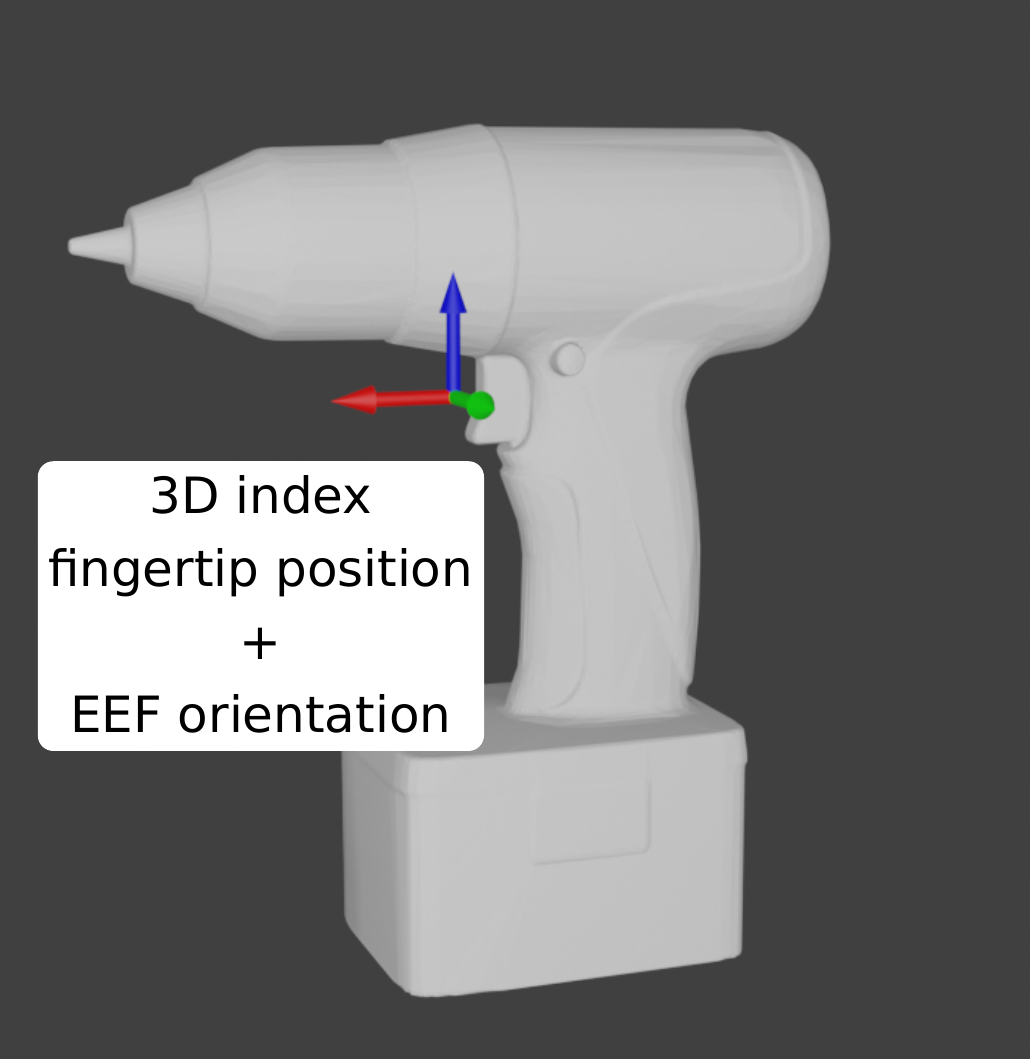}\vspace*{-1ex}
	\caption{Two target grasp representations. \emph{Left:} Explicit grasp representation, consisting of a 6D end-effector pose and finger positions. \emph{Right:} Constraint-based representation. The grasp is represented through the index fingertip 3D position and end-effector orientation. This representation allows the policy to explore different grasp configurations, satisfying the constraint.}
	\label{fig:grasp_representations}
	\vspace*{-3ex}
\end{figure}

In this section, we propose a methodology for learning a pre-grasp manipulation policy using a constraint-based target grasp representation. Specifically, we represent the target grasp as a 3D target position of the index fingertip and the end-effector rotation relative to the object. Both target grasp representations are illustrated in \reffig{fig:grasp_representations}.

The constraint-based target grasp representation offers two advantages. First, it allows the agent to explore various grasp configurations that satisfy the given target functional grasp constraint. This facilitates learning of a combination of manipulation strategies and grasp configurations. Second, it simplifies the requirements for an external oracle that has to provide the target grasp. This is because identifying key points like the trigger and the desired end-effector orientation is easier than specifying a full explicit grasp configuration.

We argue that while both grasp representations require the orientation of the object and a 3D position (object position for explicit and point of interest for constraint-based), the explicit grasp representation necessitates defining joint angles for all fingers. That creates more possibilities for errors and, hence, failed grasps. This is especially important when dealing with previously unseen object instances. Defining joint angles for each finger is more challenging than defining a 3D position of the point of interest on the object, such as a trigger of the drill, spray bottle, or handle of a mug. The downside of this constraint-based approach is a more complex learning pipeline.

\subsection{State Space}

The state representation is almost identical to the one described in \refsec{sec:State_1}. The grasp representation part is changed to reflect the constraint target grasp representation, addressed in this section. The desired functional grasp is thus provided as a column vector:
\begin{equation}
	\bm{g} = [\bm{g}^O_\textrm{ifp}, \bm{g}^O_\textrm{r}],
\end{equation}
where $\bm{g}^O_\textrm{ifp}$ is a 3D index fingertip position in the object frame of reference and $\bm{g}^O_\textrm{r}$ is a 4-element hand rotation vector represented by a quaternion. Thus, the target functional grasp is represented implicitly through a functional grasp constraint with a 7-element vector. The full state is a 52-element vector.

\subsection{Action Space}

We keep the action representation unchanged, as in \refsec{sec:Action_1}. The pose of the end-effector is controlled in 6D through IK. The finger positions are controlled directly in joint space through a coupled-joints embedding. Such action representation is generic and fits learning with the constraint-based target grasp representation well.

\subsection{Reward Function}
\label{sec:Reward_Function_2}

To reflect the semantic changes introduced by the constraint-based target grasp representation, as opposed to the explicit target grasp representation, we change the definition of a \textit{successfully achieved functional grasp} from \refeq{eq:target_1} accordingly:
\begin{equation}
	\label{eq:target_2}
	r_\textrm{T} =
	\begin{cases}
		1 & \text{if $\Delta \bm{i}_\textrm{p} < T_\textrm{p} \wedge 
			\Delta h_\textrm{r} < T_\textrm{r} \wedge 
			o_{\textrm{p}_z} > T_z$} \\
		0 & \text{otherwise,}
	\end{cases}
\end{equation}
where $\Delta \bm{i}_\textrm{p}$ is the distance from the current index fingertip position $\bm{i}_\textrm{p}$ to the desired one, $\Delta \bm{h}_\textrm{r}$ is the distance from the current end-effector rotation to the desired one, and $o_{\textrm{p}_z}$ is the Z coordinate of the object position. We use $T_\textrm{p}=1$\,cm, $T_\textrm{r}=0.15$\,rad, and $T_z=z_{\textrm{table}} + z_{\textrm{offset}}$, where $z_{\textrm{table}}$ is the height of the support plane and $z_{\textrm{offset}}=15$\,cm is an offset chosen such that any object from the dataset positioned at such height above the table cannot touch the support plane.

The last condition requires an object to be lifted off the table. This is necessary, because without it the policy could learn to satisfy the functional grasp constraint without achieving a stable grasp. For example, it is very easy to bring the index finger close to the trigger and orient the end-effector as required but have all other fingers wide open, thus not having an actual grasp. So, by accepting only grasps that enable an object to be lifted off the table, we introduce an implicit grasp stability constraint. Note that this condition is not necessary for the explicit target grasp representation. In that case, we assume that the provided explicit grasps are secure to consequently lift and use an object. This is one of the drawbacks of the explicit target grasp representation, putting more responsibility on an external oracle. In contrast, in the constraint-based grasp representation, the responsibility to ensure that the object is grasped in a stable manner is relocated to the policy.

Given the modifications from above, the \textit{reward function} from \refeq{eq:reward} becomes:
\begin{equation}
	\label{eq:reward_2}
	r(t) = r_{\textrm{grasp}}(t) +
	r_{\textrm{lift}}(t) + r_{\textrm{man}}(t) + r_{\textrm{MP}}(t) + r_\textrm{T}(t),
\end{equation}
where $r_{\textrm{grasp}}$ encourages movement towards target grasp $\bm{g}$, $r_{\textrm{man}}$ encourages pre-grasp manipulation of an object, $r_{\textrm{MP}}$ penalizes being in configurations with low manipulability, $r_\textrm{T}$ rewards reaching the target functional grasp $\bm{g}$, and the new term $r_{\textrm{lift}}$ encourages \textit{lifting an object off the table}:
\begin{equation}
	\label{eq:reward_lift}
	r_{\textrm{lift}}(t) = \min(\max((o_{\textrm{p}_z} - z_{\textrm{table}}) / z_{\textrm{offset}}, 0), 1).
\end{equation}
The lifting reward $r_{\textrm{lift}}$ is normalized $\in [0, 1]$, as are all other reward terms in this work. Not having negative rewards allows the agent to explore freely, and not having an increment-based reward (as in \refeq{eq:reward_hand}) prevents the agent from maximizing the reward by repeatedly moving an object up and down.

Finally, the \textit{grasp reward} term $r_{\textrm{grasp}}$ is modified such that:
\begin{equation}
	\label{eq:reward_grasp_2}
	r_{\textrm{grasp}} = r_{\bm{i}_\textrm{p}} + r_{\bm{h}_\textrm{r}},
\end{equation}
where $r_{\bm{i}_\textrm{p}}$ encourages moving the index fingertip position towards the target index fingertip position and $r_{\bm{h}_\textrm{r}}$ encourages moving the hand rotation towards the target grasp rotation, both being in the object frame of reference. 

The \textit{index finger reward} $r_{\bm{i}_\textrm{p}}$ is defined similar to \refeq{eq:reward_hand}:
\begin{equation}
	\label{eq:reward_hand_position_2}
	r_{\bm{i}_\textrm{p}}(t) = \frac{\Delta \bm{i}_\textrm{p}(t-1) - \Delta \bm{i}_\textrm{p}(t)}{\Delta i_\textrm{p}^\textrm{max}},
\end{equation}
where $\Delta \bm{i}_\textrm{p}$ is the Euclidean distance from the index fingertip position $\bm{i}^O_\textrm{p}$ to the target grasp index fingertip position $\bm{g}^O_\textrm{p}$. $\Delta i_\textrm{p}^\textrm{max}$ is a maximal index fingertip position change during the step duration $\Delta t$. It is computed as a sum of maximal possible end-effector velocity and maximal possible index fingertip velocity relative to a hand.

Overall, the reward function has the same structure as described in \refsec{sec:Reward_Function}, with necessary modifications to reflect a more compact and abstract target grasp representation with index fingertip position. All terms are dense and continuous, defined to be in the range $[-1, 1]$ or $[0, 1]$, so that the relative reward components scaling is straightforward.

\subsection{Curriculum} 
\label{sec:Curriculum_2}

Lifting an object above the table introduces additional complexity. To compensate for that and keep the learning time short, we extend the curriculum described in \refsec{sec:Curriculum} by introducing an additional step:
\begin{enumerate}
	\item Learning how to reach the target grasp without lifting an object. The objects are spawned upright in nominal configuration, very close to the hand. Lifting an object is excluded from the success criterion formulation.
	\item Learning how to reach the target grasp and lift an object. The hand is further away, and lifting an object is required; otherwise, it is the same as Step 1.
	\item Learning the complete objective of the pre-grasp manipulation. The objects are initialized in any combination of roll and yaw on the table.
\end{enumerate}
The three-step curriculum decomposes the learning problem in stages of gradually increasing difficulty. At the same time, it maintains a generic formulation that is straightforward to apply to arbitrary objects, arms, and hands.
\section{Evaluation}
\label{sec:Evaluation}

To evaluate the proposed approach, we apply it to the 6\,DoF UR5e robotic arm with 11\,DoF Schunk SIH hand. The joints of this wire-driven hand are coupled, leaving 5 controllable DoF. With this evaluation we try to answer the following questions:
\begin{itemize}
	\item Does our approach reliably produce robust manipulation policies, capable of dexterous pre-grasp manipulation of unseen object instances of a known category?
	\item Does the multi-component manipulation reward $r_\textrm{man}$ lead to policies with higher success rates?
	\item Does the curriculum improve convergence stability?
	\item Does our approach enable learning of feasible manipulation policies for both target grasp representations?
\end{itemize}

\subsection{Setup}

We use Proximal Policy Optimization (PPO)~\cite{Schulman_2017} to train the policies. We employ the RL Games~\cite{rl_games_2021} high-performance implementation for GPU parallelization. We use the findings of Mosbach et al.~\cite{Mosbach_2022} as a base, keeping the learning algorithm hyperparameters the same. The policy is represented by a three-layer, fully-connected neural network. In our case, the input is a 57-element vector. The network is a multilayer perceptron and has the following structure:\\ \hspace*{5ex}\mbox{$57 \times 512 \rightarrow 512 \times 256 \rightarrow 256 \times 128 \rightarrow 128 \times 11$}.

\begin{figure}[t]
	\centering
	\includegraphics[width=0.32\linewidth]{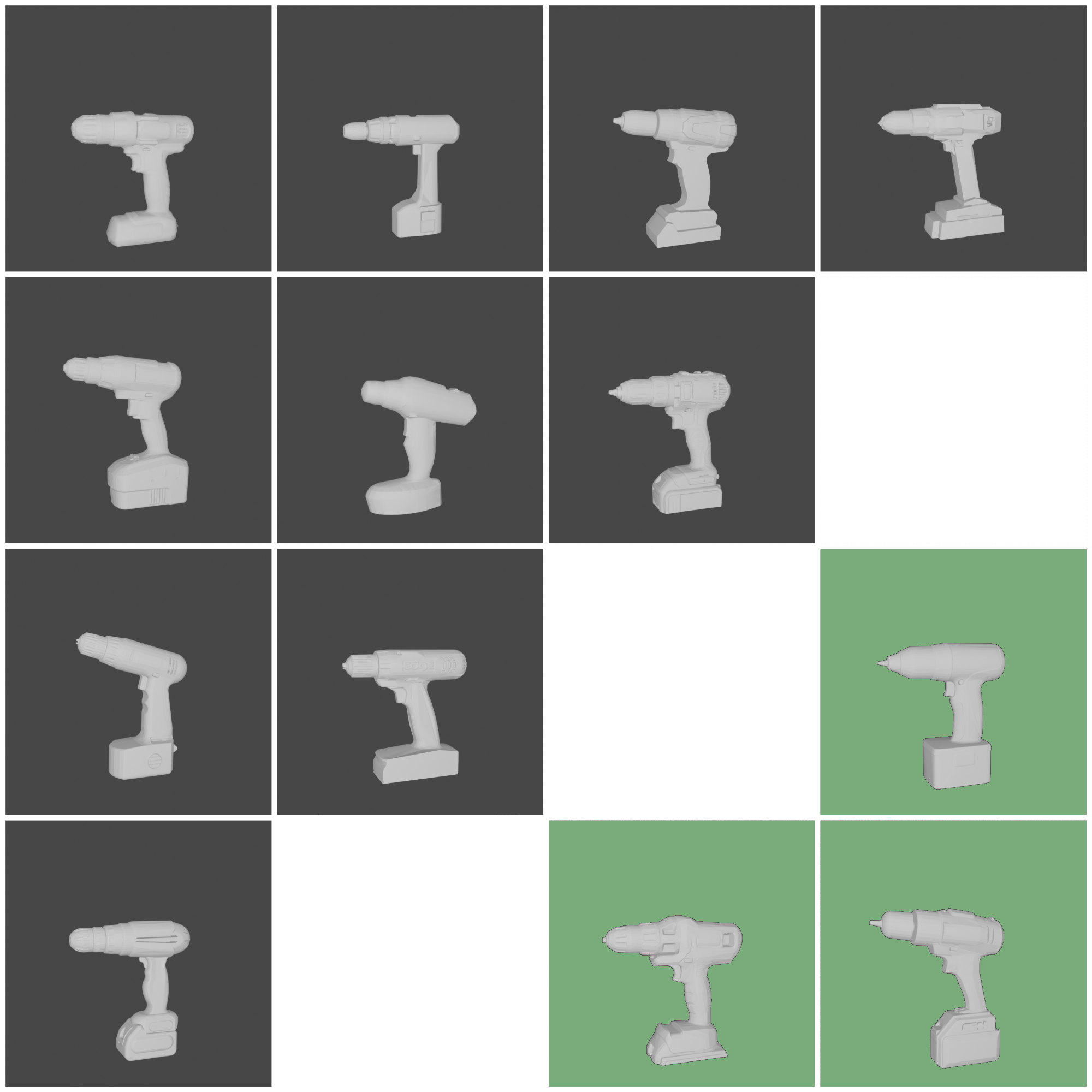}
	\includegraphics[width=0.32\linewidth]{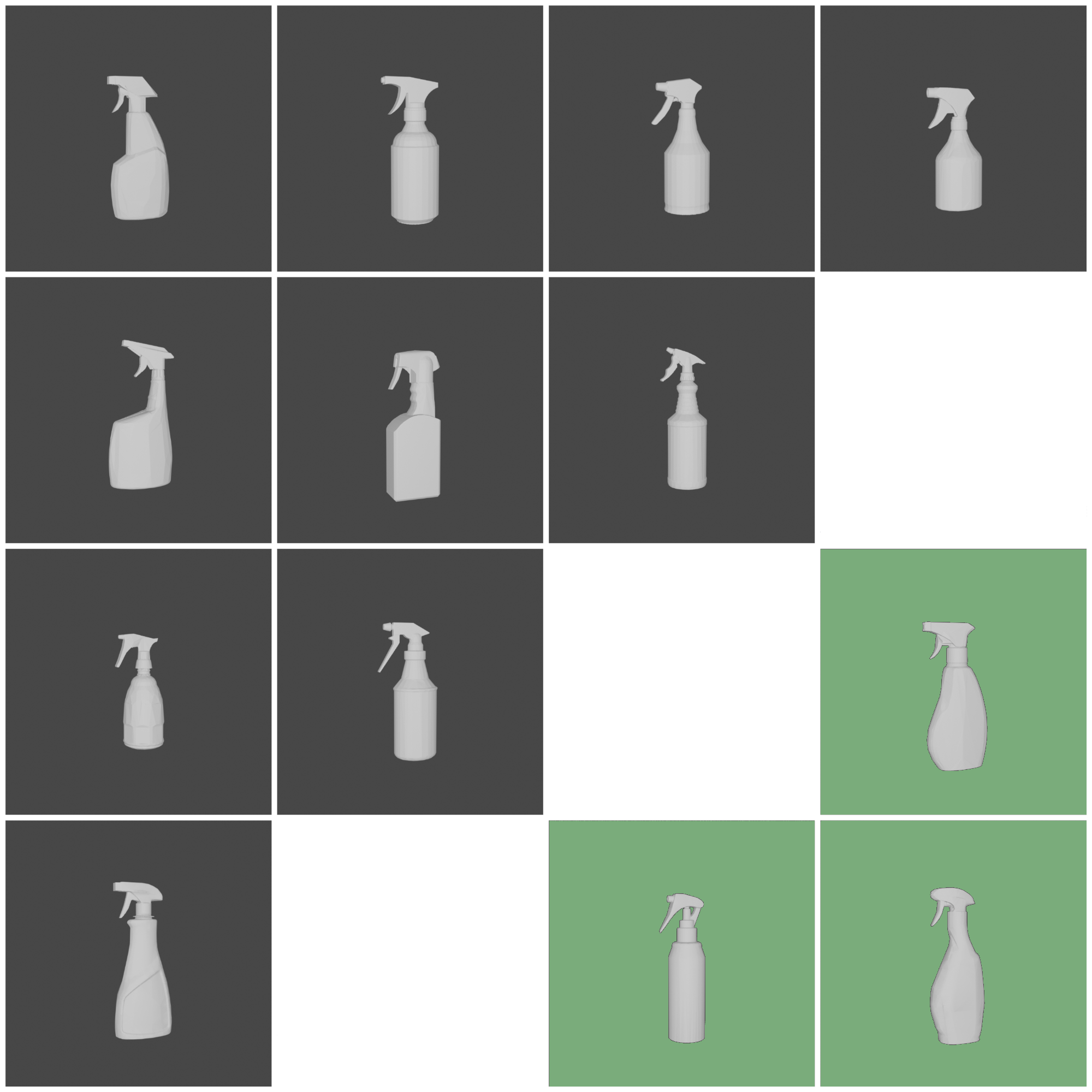}
	\includegraphics[width=0.32\linewidth]{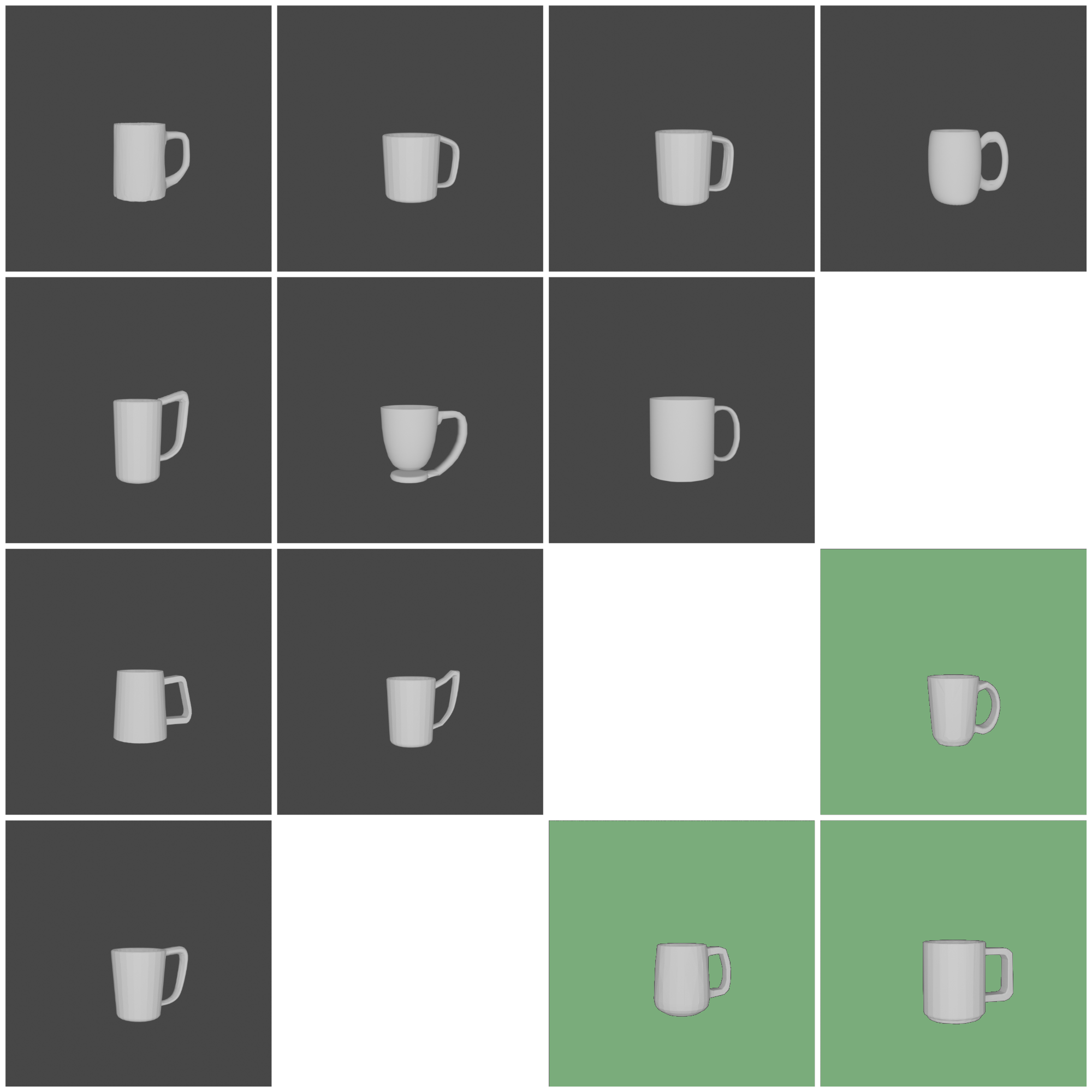}
	\caption{Dataset of 39 objects of three categories: drills, spray bottles, and mugs. Each category has 13 objects: ten for training (gray background) and three for testing (green background).}
	\label{fig:dataset}
	\vspace*{-3ex}
\end{figure}

In our experiments, we pursue the objective of learning a single functional grasping policy for three rigid object categories: drills, spray bottles, and mugs. To this end, we prepared a 3D mesh dataset of 39 objects: 13 of each category, where ten objects are for training and the remaining three objects are used for testing. The dataset is shown in \reffig{fig:dataset}. It is composed of meshes from~\cite{Rodriguez_2018a} and of meshes available online\footnote{\url{https://free3d.com}, \url{https://3dsky.org}}. We make the dataset available online\footnote{\url{https://github.com/AIS-Bonn/fun_cat_grasp_dataset}}.

We select these three specific object categories as they represent objects that are functionally grasped in three different ways relative to their Center of Mass~(CoM). Drills are functionally grasped roughly at CoM. Spray bottles are functionally grasped far from CoM along the Z axis of the object. Mugs are grasped far from CoM along the X-axis of the object. At the same time, these categories also have diverse shape properties. Drills have complex shapes with multiple graspable regions. Spray bottles are elongated, making them difficult to position upright without falling. At the same time, some of them feature smooth cylindrical shapes that can be easily rolled, while the others resemble parallelepiped-like shapes and cannot be easily rolled. Finally, mugs consist of one large cavity for holding liquid and a smaller, through-going space formed by the inner area of the handle. These fundamental differences in functional grasp regions relative to CoM and in overall shape make these three categories suitable candidates to form a compact but diverse dataset, requiring a certain level of generalization due to the high variability in optimal strategies.

To estimate the variance within the dataset, we compared each instance of the binary projections onto 2D (with the viewpoint as shown in \reffig{fig:dataset}) to the corresponding median 2D shape within each category. The average variance is 0.41 for drills, 0.32 for spray bottles, and 0.45 for mugs. The variance values were normalized within each category so that 0 is the least varying instance and 1 is the most varying instance.

We use the high-performance GPU physics simulator Isaac Gym~\cite{Makoviychuk_2021}. The experiments are performed on a single NVIDIA RTX A6000 GPU with 48\,GB of VRAM.

In this work, we assume that the objects are located on a table in front of the robot. Thus, there are three possible natural poses in which drills, spray bottles, or mugs can be: standing upright and lying on their left or right side. All other possible poses on a flat surface are unstable and transition quickly to one of the described poses. Mugs can also be positioned upside down. However, we do not use this pose in our experiments to ensure that the results are consistent and comparable between all object categories.

Actions are generated with a frequency of 30\,Hz. The objects are spawned on a table in front of the robot, such that at least 75\% of their bounding box is in the manipulation workspace. Poses in which objects are lying on their sides are the most challenging for functional grasping because of the occlusion.
For this reason, we focus on such poses and use the following object rotation distribution: 20\% of the objects are upright, 40\% are on their left side, and 40\% are on their right side. The yaw angle and the object position are sampled uniformly. The hand starts at a random 6D pose above the table. Notably, objects lying on the right side are more challenging to functionally grasp with the right hand. Learning is performed on the training set of 30 objects. A target functional pre-grasp is defined for each object manually.

To make the simulation setup more realistic, Gaussian noise is applied to all observations supplied to the policy. For positions and distances, the zero-mean noise has $\sigma=3$\,mm. For rotations, the zero-mean noise has $\sigma=5^{\circ}$. Before each action is generated, the noise values are drawn from the distribution specified above and added to the ground truth values from the simulation. The only two observations that do not have noise are the object category and the target grasp. In each environment, an object is assigned a realistic random mass. The mass distribution in kg per category is represented by a Gaussian: $\mathcal N(1.4, 0.2)$ for drills, $\mathcal N(0.5, 0.15)$ for spray bottles, and $\mathcal N(0.3, 0.07)$ for mugs. Both the noise and the mass are limited to deviate from the mean for not more than $3\sigma$. The new mass values are set before each episode.

We keep the reward value for reaching the target grasp $r_{T} = 5,000$, as it is the default value in the RL Games framework. We scale the reward components: orienting reward $r_\textrm{orient}$ by 500 and holding reward $r_\textrm{hold}$ by 25. The other reward components are not scaled. Both scaling factors are chosen to be one order of magnitude less than the final reward and the corresponding reward component with higher scaling.

The order of the scaling is defined by the desired action sequence: first the hand approaches the object (scale 1), then it gains control over the object by holding it (scale 25), then it orients the object (scale 500), and finally, after repositioning the object, the target grasp is achievable (scale 5,000). Switching the order of scaling often hinders the progress towards learning of achieving the target grasp. That happens because the policy is greedily maximizing reward through, for example, holding an object, rather than exploring more difficult and failure-prone orienting the object if it were yielding less reward. The exact values of these scaling factors do not affect learning significantly, as long as the overall proportions reflect the logical sequence: approach, hold, orient, grasp.

\begin{figure}[t]
	\centering
	\includegraphics[width=7cm]{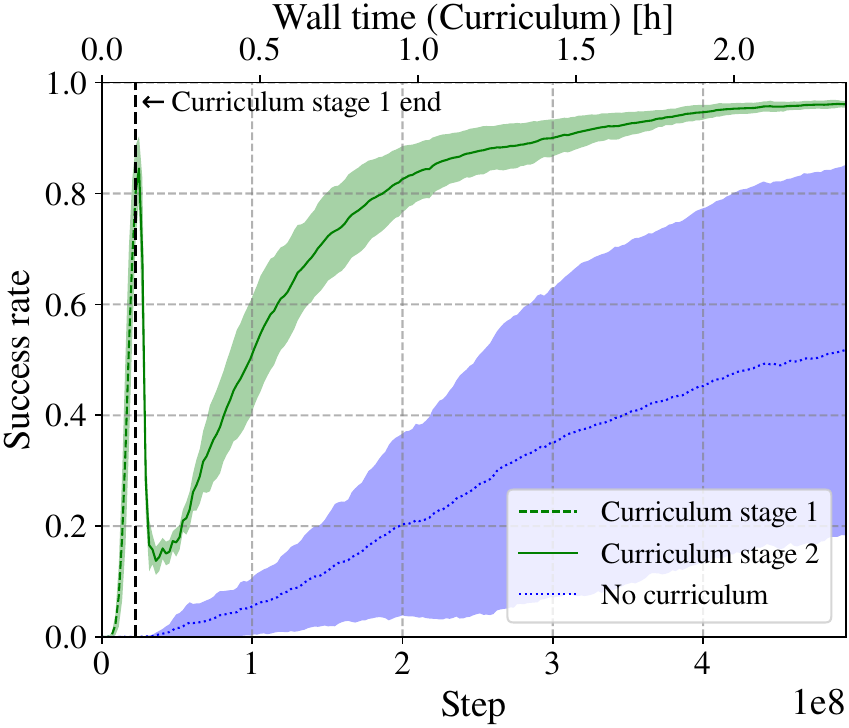}\vspace*{-1ex}
	\caption{Explicit target grasp representation: Curriculum ablation experiment training curves. The two-stage curriculum significantly improves convergence stability compared to the runs without the curriculum. Lines: means. Colored areas: 95\% confidence intervals.}
	\label{fig:sr_curriculum}
	\vspace*{-3ex}
\end{figure}

\subsection{Explicit Target Grasp Representation}
\label{sec:experiments_1}

In this section, we evaluate the approach using an explicit target grasp representation that is described in \refsec{sec:Method}. We train the policy on a single GPU with 16,384 parallel environments. Each policy in this evaluation is trained three times with three different seeds to assess convergence stability.

An episode terminates when (i) a target functional pre-grasp is reached, (ii) an object falls from the table, or (iii) a maximum number of steps is reached. We set the maximum number of steps to 200, which corresponds to $\approx\!6.7$ seconds. We assume that provided explicit target pre-grasps are valid and enable lifting and manipulating an object; thus we do not require the policy to lift objects off the table once the grasps are achieved during the training stage.

First, we perform an ablation study of the two-stage curriculum proposed in \refsec{sec:Curriculum}. During the first stage, the objects are spawned with a nominal rotation and close to the hand. Since at this stage the grasp is easily reachable, we disable the manipulation reward $r_\textrm{man}$ to ensure quicker convergence. The first stage continues until at least a 50\% success rate is achieved for each object instance. During the second stage, the objects are spawned with the rotations described above, and the full reward is used.

\reffig{fig:sr_curriculum} shows learning curves during learning with and without curriculum. In addition, the wall-clock time is shown for the curriculum runs. The wall time of other runs is similar ($\pm 10$\,min). The first stage of the curriculum is completed quickly. The second stage takes longer, but all three runs reliably converge to a success rate of 97\% in under three hours with little variance. Without the curriculum, the policy achieves only $\approx\!50$\% success rate and has a large variance within runs.
Hence, the two-stage curriculum significantly improves convergence stability and success rate. 

We use a default discounting factor $\gamma = 0.95$ in all experiments. Lower values, such as $\gamma = 0.9$, decrease learning speed significantly, since only short time spans are represented in the rewards, leaving longer-term consequences of complex manipulation unrepresented. Higher values, such as $\gamma = 0.975$, do not provide any significant improvement.

\begin{figure}[t]
	\centering
	\includegraphics[width=7cm]{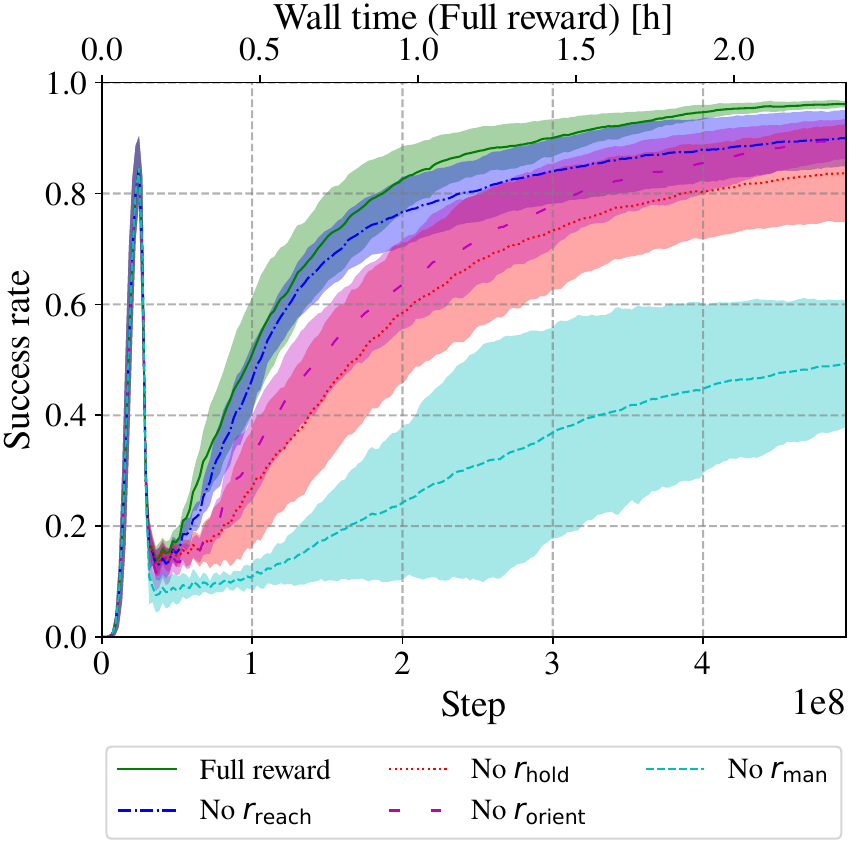}
	\caption{Explicit target grasp representation: Manipulation reward ablation experiment training curves. Disabling single reward components slightly deteriorates  convergence rate and stability. Disabling the whole manipulation reward component makes the learning process significantly slower and less stable. Lines: means. Colored areas: 95\% confidence intervals.}
	\label{fig:sr_ablation}
	\vspace*{-3ex}
\end{figure}

\begin{figure*}[t]
	\centering
	\includegraphics[width=0.155\linewidth]{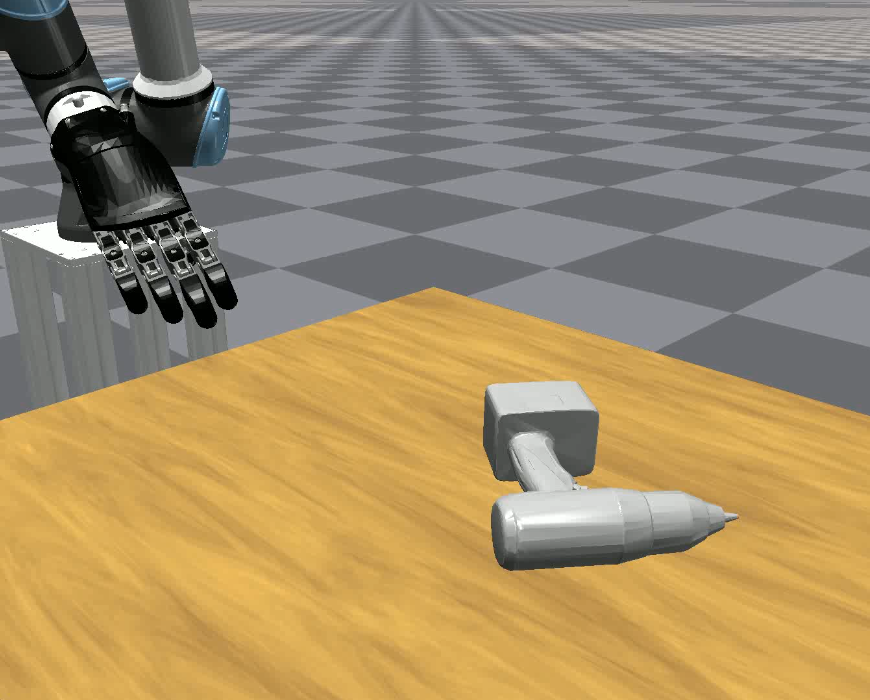}\hspace{0.5ex}
	\includegraphics[width=0.155\linewidth]{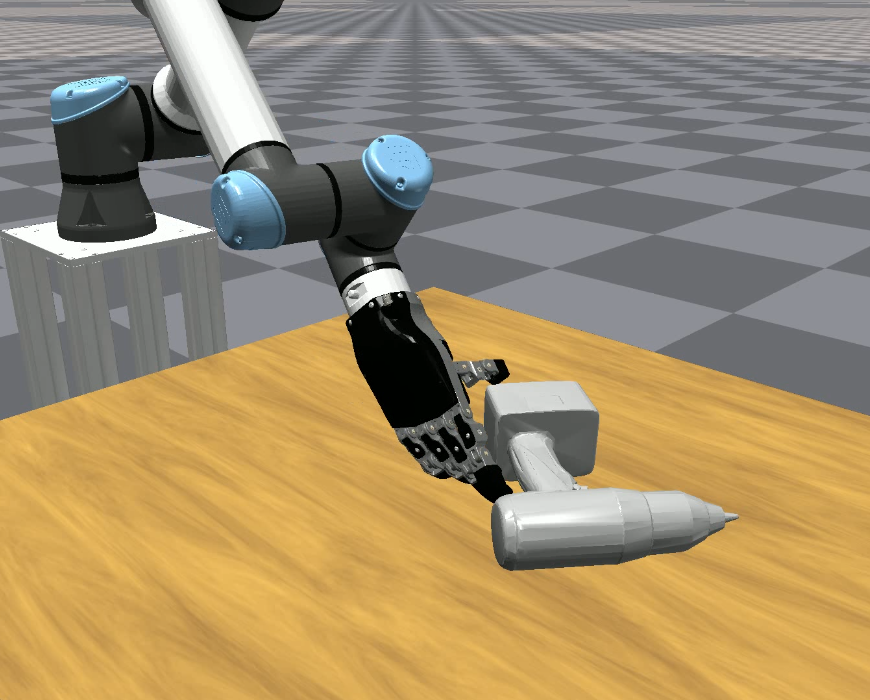}\hspace{0.5ex}
	\includegraphics[width=0.155\linewidth]{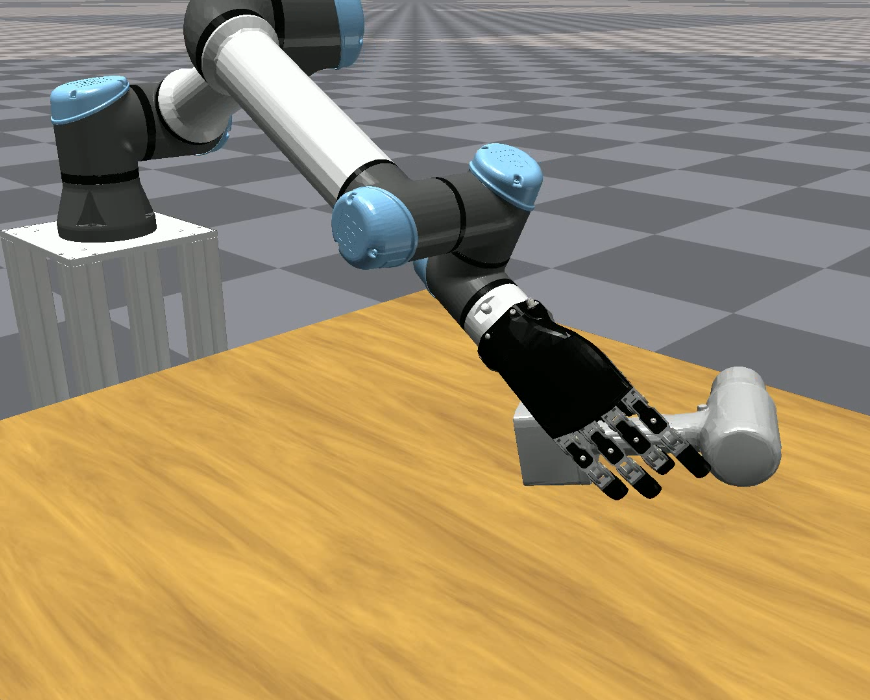}\hspace{0.5ex}
	\includegraphics[width=0.155\linewidth]{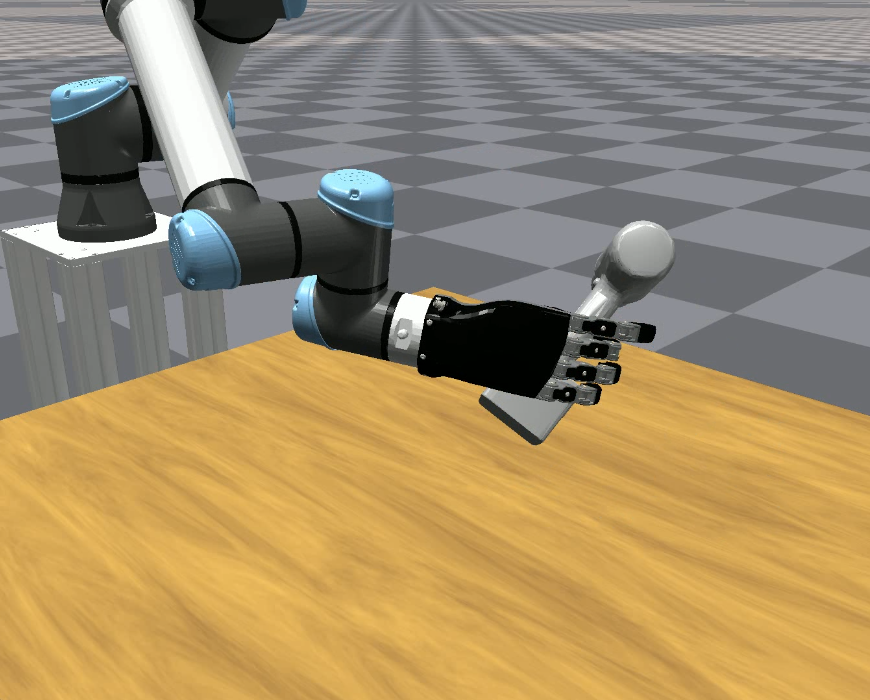}\hspace{0.5ex}
	\includegraphics[width=0.155\linewidth]{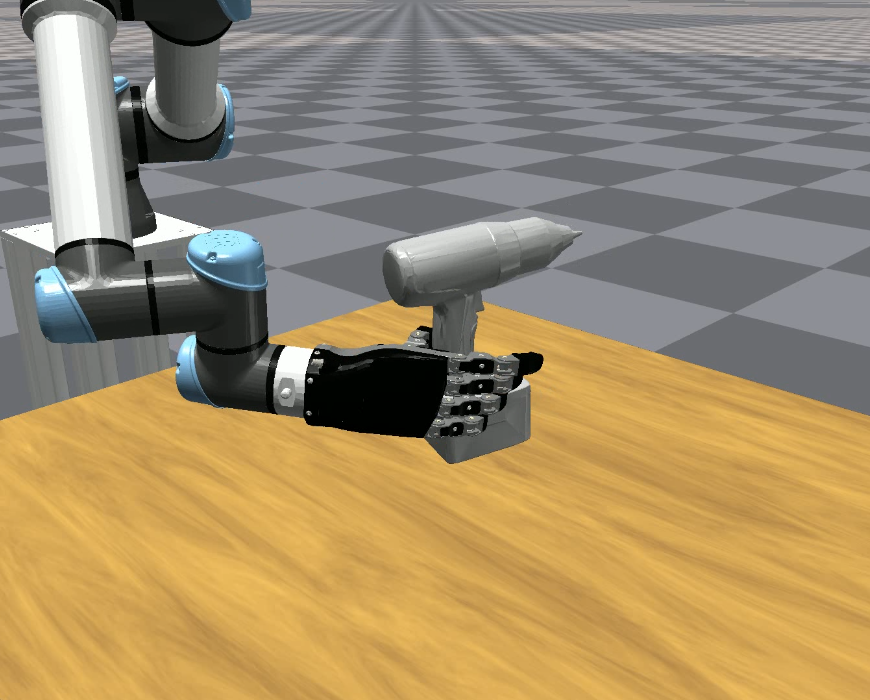}\hspace{0.5ex}
	\includegraphics[width=0.155\linewidth]{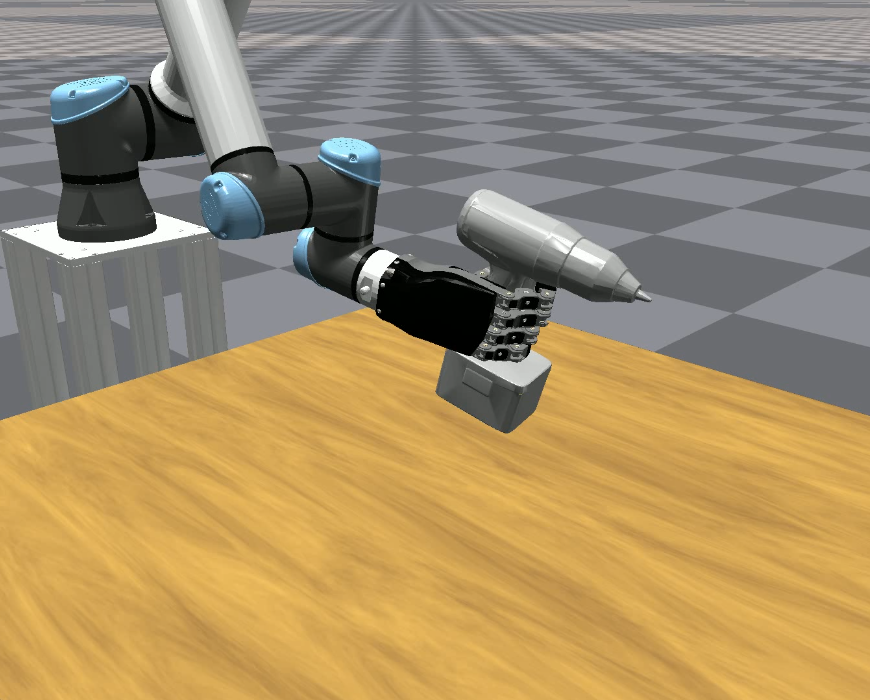}
	\\ \vspace*{1ex}
	\includegraphics[width=0.155\linewidth]{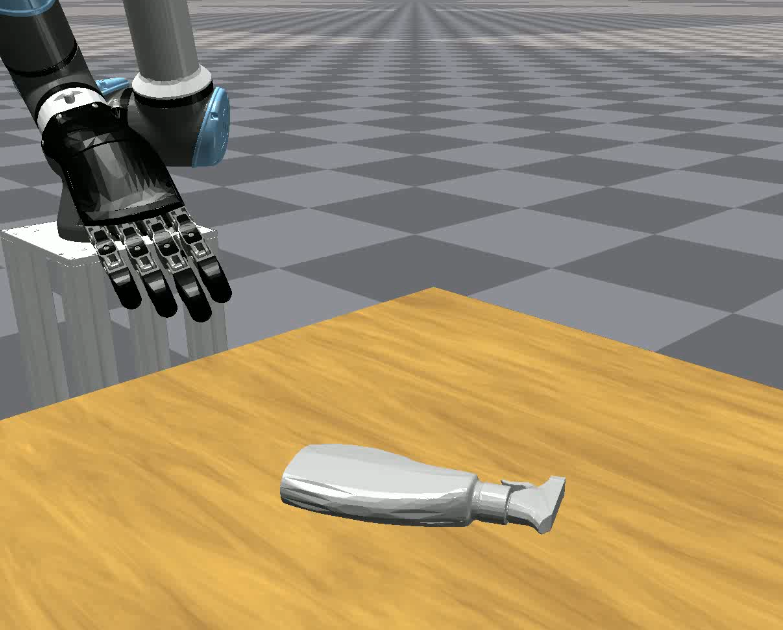}\hspace{0.5ex}
	\includegraphics[width=0.155\linewidth]{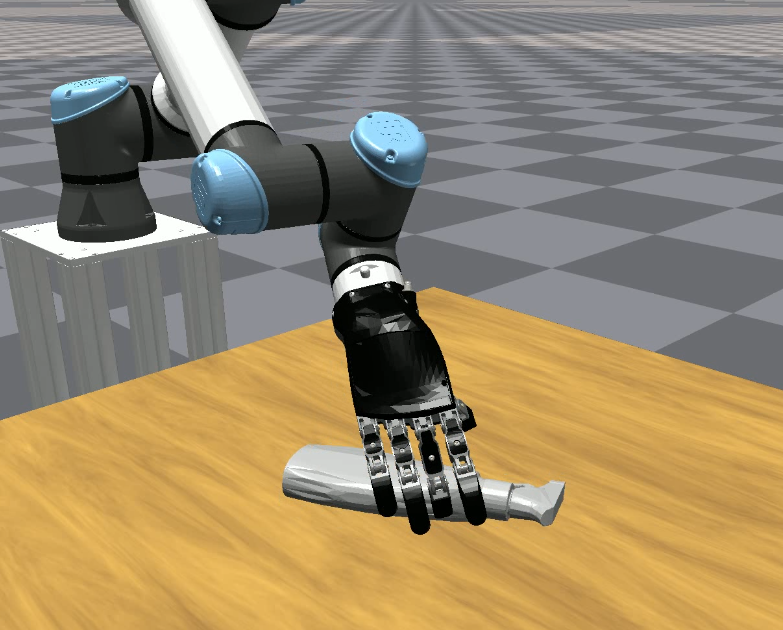}\hspace{0.5ex}
	\includegraphics[width=0.155\linewidth]{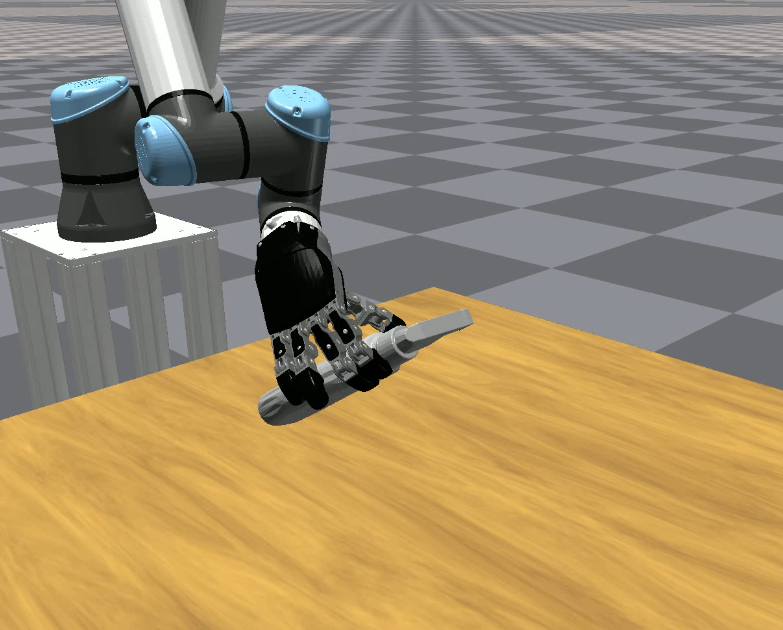}\hspace{0.5ex}
	\includegraphics[width=0.155\linewidth]{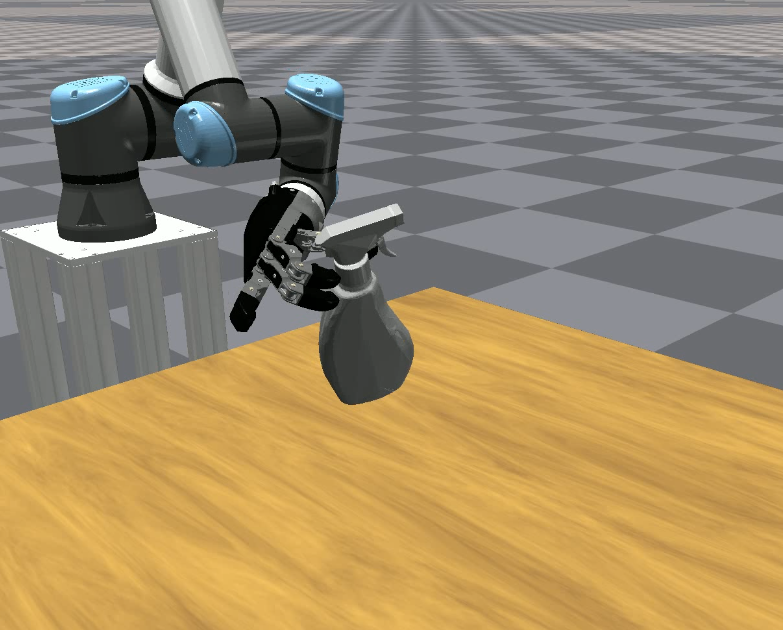}\hspace{0.5ex}
	\includegraphics[width=0.155\linewidth]{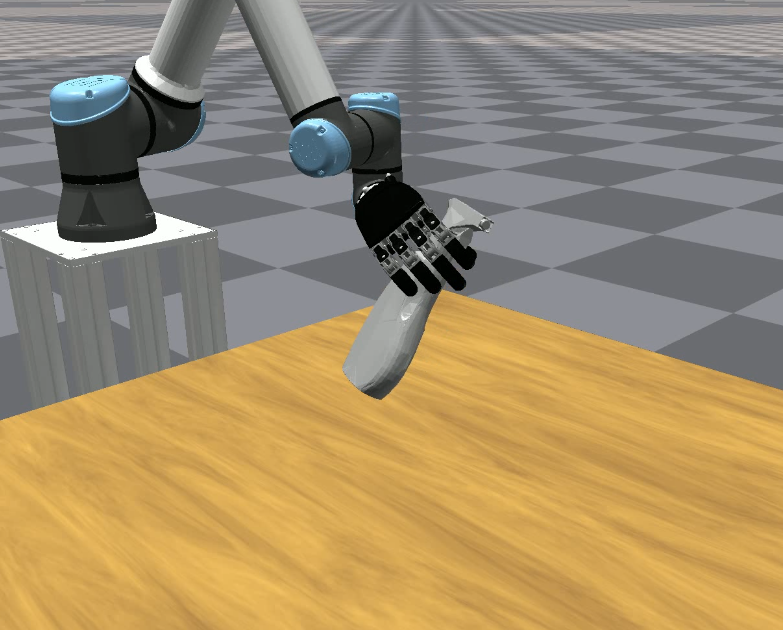}\hspace{0.5ex}
	\includegraphics[width=0.155\linewidth]{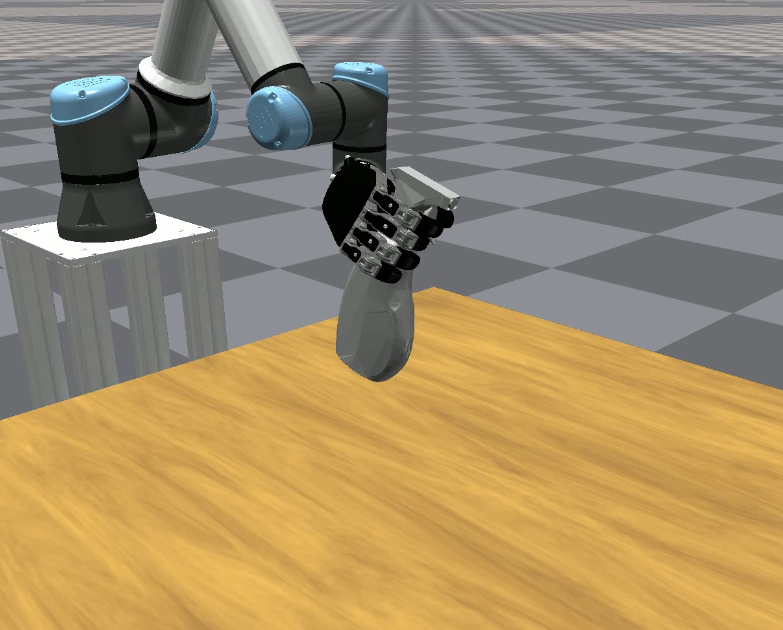}
	\\ \vspace*{1ex}
	\includegraphics[width=0.155\linewidth]{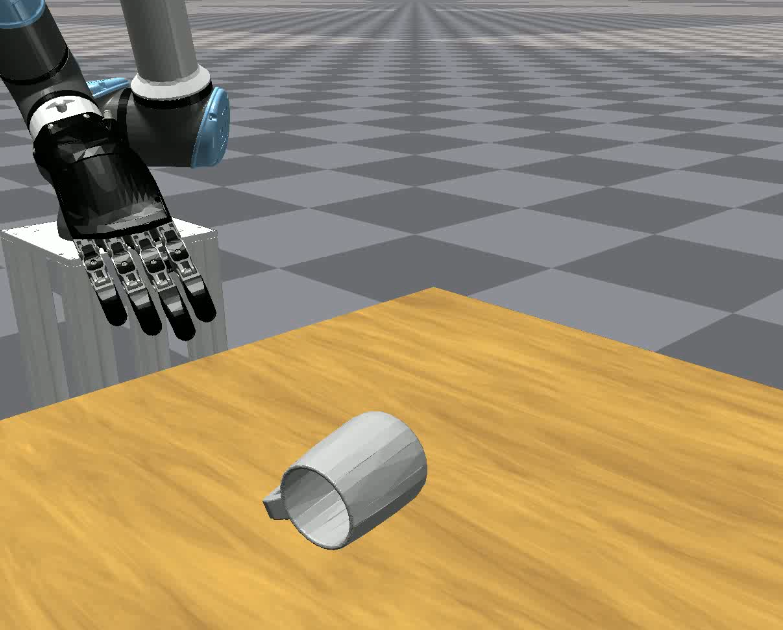}\hspace{0.5ex}
	\includegraphics[width=0.155\linewidth]{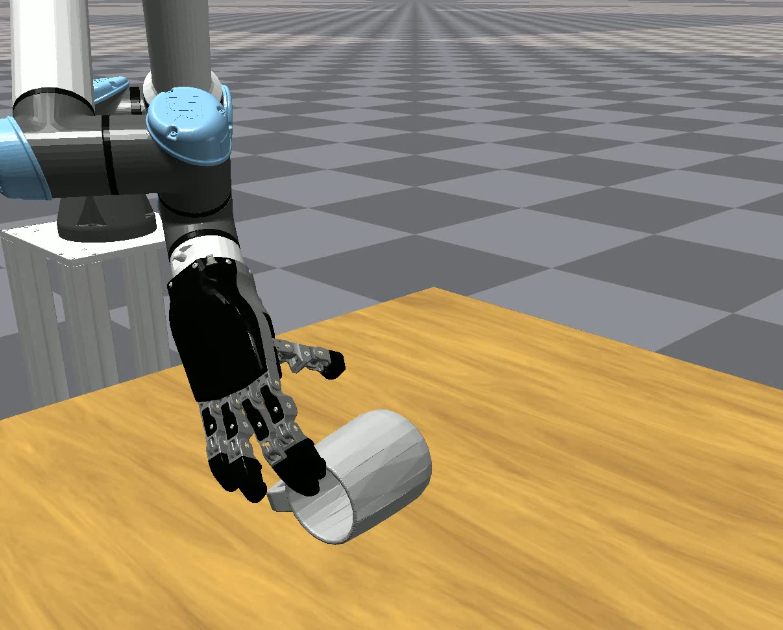}\hspace{0.5ex}
	\includegraphics[width=0.155\linewidth]{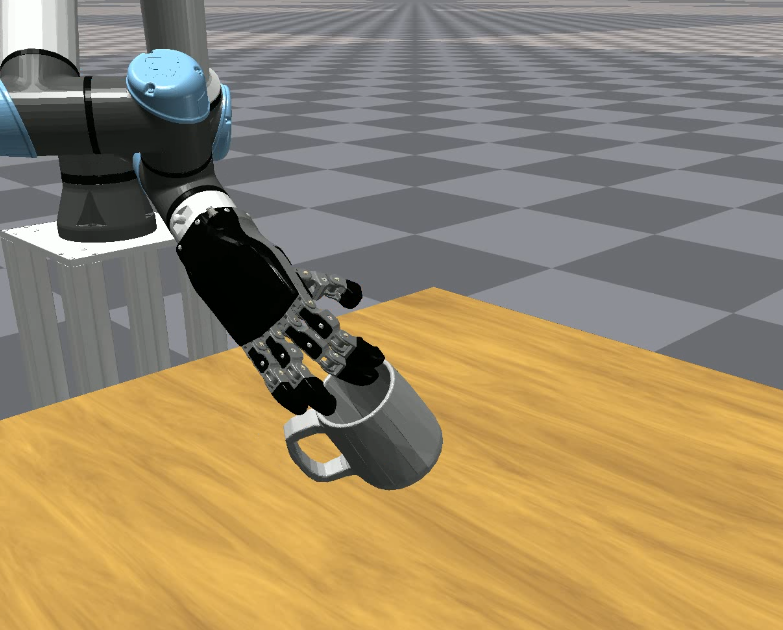}\hspace{0.5ex}
	\includegraphics[width=0.155\linewidth]{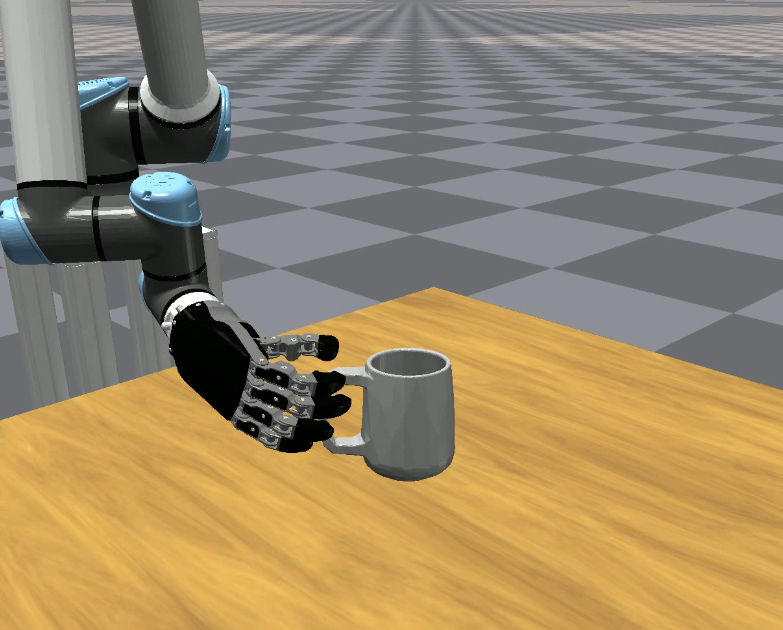}\hspace{0.5ex}
	\includegraphics[width=0.155\linewidth]{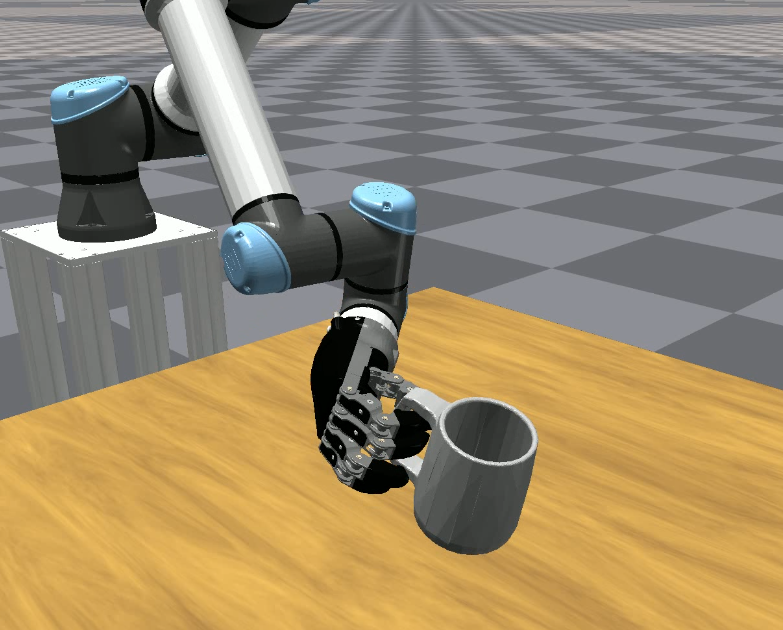}\hspace{0.5ex}
	\includegraphics[width=0.155\linewidth]{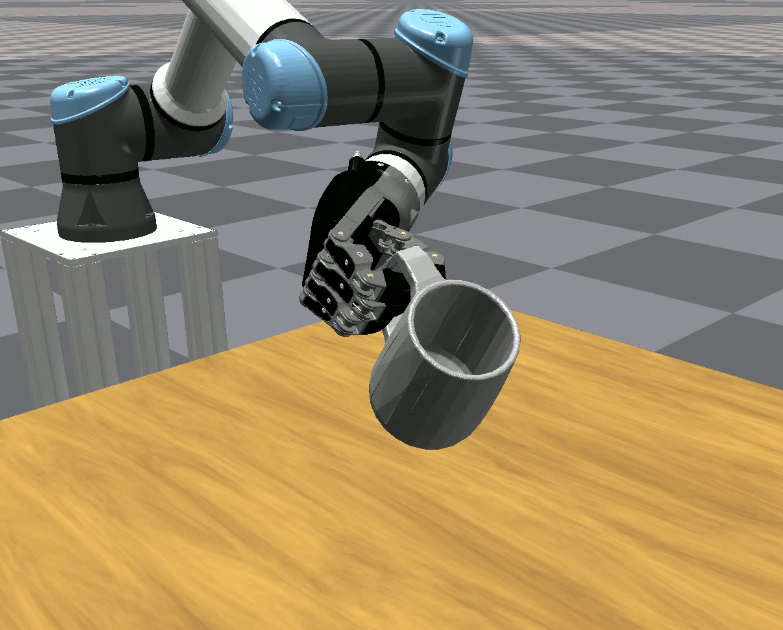}
	\caption{Explicit target grasp representation: Rollouts of learned policy manipulating unseen objects of known categories, positioned in a way that a direct functional grasp is impossible. Top to bottom: drill, spray bottle, and mug. Note the functional grasps achieved in the end.}
	\label{fig:rollouts}
	\vspace*{-3ex}
\end{figure*} 

Next, we conduct an ablation study of the proposed multi-component reward function. We train five policy variants: (i) full reward; (ii) with the reward component $r_\textrm{reach}$ encouraging moving the hand towards the object disabled;  (iii) with reward component $r_\textrm{hold}$ encouraging holding the object disabled; (iv) with the reward component $r_\textrm{orient}$ encouraging rotating the object towards the nominal rotation  disabled; and finally (v) with the whole manipulation reward component $r_\textrm{man} = r_\textrm{reach} + r_\textrm{hold} + r_\textrm{orient}$ disabled.

\reffig{fig:sr_ablation} shows the learning curves for this ablation study. One can observe that when a single component of the manipulation reward is disabled, the policy learns to achieve the goal slower but still reliably makes progress towards a high success rate. The most important component is $r_\textrm{hold}$. Without $r_\textrm{hold}$, the policy has the highest variance within runs and achieves the lowest success rate among single-component ablations. This shows that encouraging holding behavior is essential.

The deteriorated but still reliable convergence without single reward terms suggests that although each component is important, the formulation is generic enough to not depend on every detail. In contrast, disabling the whole manipulation reward $r_\textrm{man}$ has a drastic negative effect on the performance of the policy. Although it achieves a success rate of 50\%, it struggles to learn a robust behavior for objects in difficult configurations. Overall, this ablation study demonstrates that the proposed manipulation reward component significantly speeds up the learning of dexterous pre-grasp manipulation.

\begin{table}[b]
	\vspace*{-2ex}
	\centering
	\caption{Success rates with explicit grasp representation}
	\label{table:success_rate}
	\normalsize
	\begin{tabular}{lcc}
	\hlinewd{1pt} 
		Category      & Training set    & Test set     \\ \hlinewd{1pt}
		Drills        & 96.0 $\pm$ 1.2 & 94.3 $\pm$ 2.6 \\ \hline
		Spray bottles & 97.7 $\pm$ 0.9 & 92.3 $\pm$ 3.0 \\ \hline
		Mugs          & 99.3 $\pm$ 0.5 & 95.6 $\pm$ 2.3 \\ \hline \hline
		All three     & 97.7 $\pm$ 0.5 & 94.1 $\pm$ 1.5 \\ \hlinewd{1pt}
	\end{tabular}
	\begin{tablenotes}
		\item \hspace{2ex} \footnotesize{Success rates in \%. Mean $\pm$ 95\% confidence interval.}
	\end{tablenotes}
\end{table}

To assess the generalization capabilities of the learned policy, we measure its success rate, both on the training set and the test set. We use the policy trained with curriculum and the full reward. The training set consists of 30 objects, ten for each of the three categories. The test set consists of nine novel objects of the known categories. We perform 100 grasping attempts for each object. This results in 3,000 attempts for the training set and 900 attempts for the test set.

Object initial rotations are sampled as during learning: 20\% upright, 40\% on the left side, and 40\% on the right side. Once the target pre-grasp is reached, the success is tested by closing the hand. If the object stays in the hand and the key condition of a functional grasp, such as an index finger on the trigger, is satisfied, an attempt is considered successful. We allocate 300 steps or 10\,s per episode.

The measured success rates for all object categories are reported in \reftab{table:success_rate}. On the training set, the learned policy shows a high success rate of 97.7\%. As expected, on the test set the success rate is lower, but still high at 94.1\%. The highest success rates are achieved on mugs. This is because they are relatively easy to flip over from the side position and have a simple geometry. The hardest object category is the spray bottles. This is because spray bottles are narrow, have a high CoM, and can be easily dropped. 

\reffig{fig:rollouts} shows example rollouts for three test set objects. One can observe that dexterous interactive pre-grasp manipulation has been learned that leads to functional grasps for all three object categories. Videos of the learned interactive functional grasping behavior are available online\footnote{\url{https://www.ais.uni-bonn.de/videos/TASE_2024_Pavlichenko}}. One can observe the complex pre-grasping strategies such as repositioning the object, reorienting and uprighting the object, and regrasping are executed. The policy learned to reattempt the sub-tasks in case of failures.

\subsection{Constraint-based Target Grasp Representation}
\label{sec:experiments_2}

In this section, we evaluate the approach using the more abstract constraint-based target grasp representation presented in \refsec{sec:Method_2}. The policy is represented with the same model as in the experiments with explicit target grasp representation. Training procedure, simulation setup, and hyperparameters are identical as well. An episode is terminated when (i) a provided target constraint---defining the functional grasp---is satisfied and the object is lifted off the table, (ii) an object falls from the table, or (iii) a maximum number of 200 steps is reached.

\begin{figure}[t]
	\centering
	\includegraphics[width=7cm]{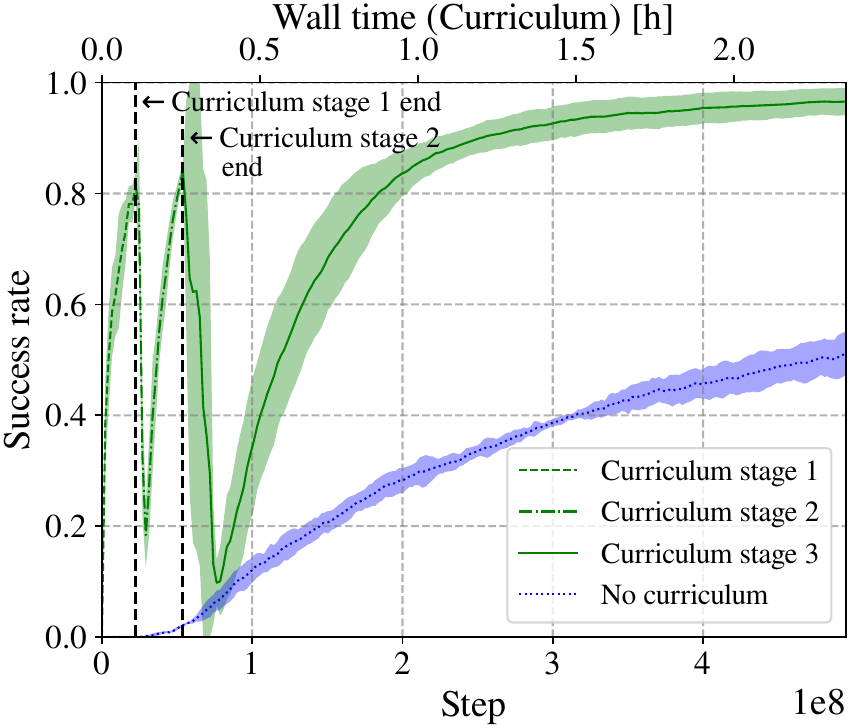}\vspace*{-1ex}
	\caption{Constraint-based target grasp representation: Curriculum ablation experiment training curves. The three-stage curriculum significantly improves learning speed. Lines: means. Colored areas: 95\% confidence intervals.}
	\label{fig:sr_curriculum_constraint}
	\vspace*{-3ex}
\end{figure}

Learning curves, averaged over three training runs, are shown in \reffig{fig:sr_curriculum_constraint}. The training is done according to the three-stage curriculum described in \refsec{sec:Curriculum_2}. The first stage is dedicated to learning to satisfy given functional grasp constraints while the objects are in easily accessible configurations. The second stage requires the policy to lift the objects off the table. With this stage, we implicitly enforce the policy to learn reliable grasp configurations. At the third stage, the objects are positioned in configurations where a direct functional grasp cannot be reached. We also perform runs without the curriculum to explore its importance.

One can see that with the curriculum, the policy reliably converges to the success rate of around $\approx\!93$\% in under three hours. The first two stages of the curriculum are completed quickly. In contrast, without the curriculum, the policy reaches only a subpar success rate of $\approx\!50$\%. It is worth noting that between runs without the curriculum there is not much variance, compared to the explicit target grasp representation runs shown in \reffig{fig:sr_curriculum}. We attribute this effect to the higher difficulty of the learning problem for the constraint-based target grasp representation since the policy has to find a way to grasp the objects to enable successful lifting on its own. Given that, the probability of having a lucky run where the policy learns how to manipulate and grasp all objects without the curriculum is much lower compared to runs with an explicit target grasp representation.

In addition, we also perform an ablation study of the proposed multi-component reward function in the context of the constrained target grasp representation. We train five policy variants: (i) with full reward; (ii) with reward component $r_\textrm{reach}$ encouraging moving the hand towards the object disabled; (iii) with reward component  $r_\textrm{hold}$ encouraging holding the object disabled;  (iv) with reward component $r_\textrm{orient}$ encouraging rotating the object towards the nominal rotation disabled; and (v) with whole manipulation component $r_\textrm{man} = r_\textrm{reach} + r_\textrm{hold} + r_\textrm{orient}$ disabled.

\begin{figure}[t]
	\centering
	\includegraphics[width=7cm]{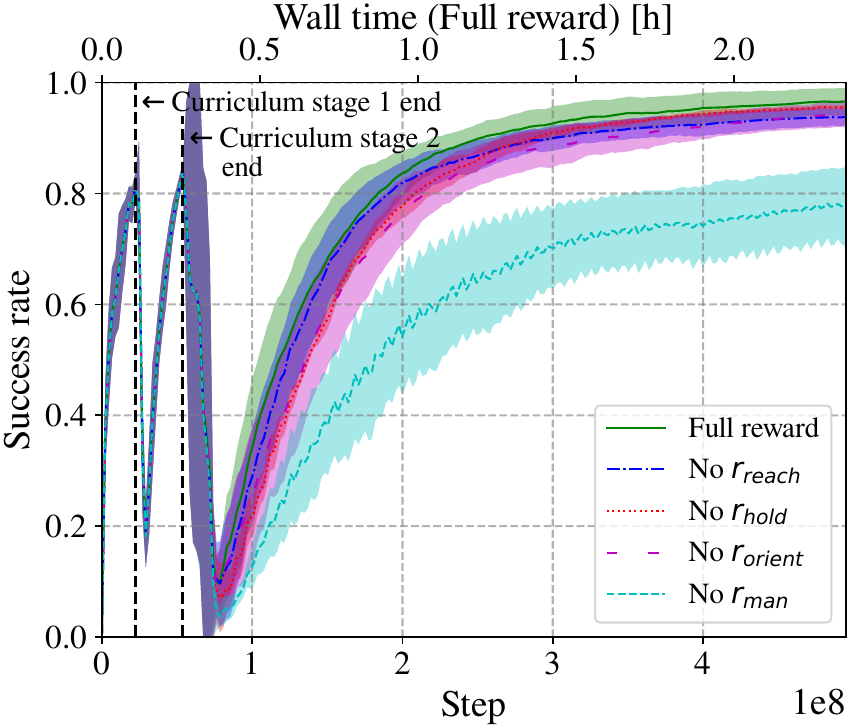}
	\caption{Constraint-based target grasp representation: Manipulation reward ablation experiment training curves. Disabling single reward components slightly deteriorates the convergence rate and stability. Disabling the whole manipulation reward component makes the learning process slower and less stable. Lines: means. Colored areas: 95\% confidence intervals.}
	\label{fig:sr_ablation_constraint}
	\vspace*{-3ex}
\end{figure}

\reffig{fig:sr_ablation_constraint} shows the learning curves for this ablation study. One can see that similar to \reffig{fig:sr_ablation}, all ablation runs yield worse success rates compared to the policies with full reward. However, the difference is less significant compared to the explicit grasp representation ablation. While disabling the whole manipulation reward component significantly decreased learning speed, disabling individual components affects the learning performance insignificantly. We attribute this to the fact that in the case of the constrained target grasp representation, the policy has to figure out the way to grasp objects by itself, forced by the requirement to lift the objects. While doing that, the policy implicitly acquires holding behaviors. Nevertheless, the ablation study demonstrated that the proposed manipulation reward facilitates quicker and more stable convergence for policies learned with the explicit grasp representation.

\begin{table}[!b]
	\vspace*{-2ex}
	\centering
	\caption{Success rates with constraint-based grasp representation}
	\label{table:success_rate_constraint_based}
	\normalsize
	\begin{tabular}{lcc}
		\hlinewd{1pt} 
		Category      & Training set    & Test set     \\ \hlinewd{1pt}
		Drills        & 92.7 $\pm$ 1.7 & 90.3 $\pm$ 3.3 \\ \hline
		Spray bottles & 93.8 $\pm$ 1.7 & 88.6 $\pm$ 3.5 \\ \hline
		Mugs          & 92.0 $\pm$ 1.8 & 91.1 $\pm$ 3.1 \\ \hline \hline
		All three     & 92.8 $\pm$ 1.0 & 90.1 $\pm$ 1.9 \\ \hlinewd{1pt}
	\end{tabular}
	\begin{tablenotes}
		\item \hspace{2ex} \footnotesize{Success rates in \%. Mean $\pm$ 95\% confidence interval.}
	\end{tablenotes}
\end{table}

To quantitatively evaluate the learned policy, we perform 100 attempts for each object from the training set and for novel instances of known categories in the test set, the same as in \refsec{sec:experiments_1}. This results in 3,000 attempts for the training set and 900 attempts for the test set.

The resulting success rates are reported in \reftab{table:success_rate_constraint_based}. As expected, the policy has a higher success rate on the training set. On the novel instances from the test set, the policy achieves a 90\% success rate, which is slightly lower than 94\% observed for the explicit target grasp representation (\reftab{table:success_rate}). We attribute the lower performance both to a more difficult task, which includes lifting the object, and to a more abstract grasp representation. On the other hand, the constraint-based target grasp representation has fewer requirements for an external oracle, compared to the policy with explicit target grasp representation. Specifically, in our experiments we had to put much effort into specifying the explicit target grasps, while defining the target constraints is quick and straightforward.

\begin{figure*}[t]
	\centering
	\includegraphics[width=0.155\linewidth]{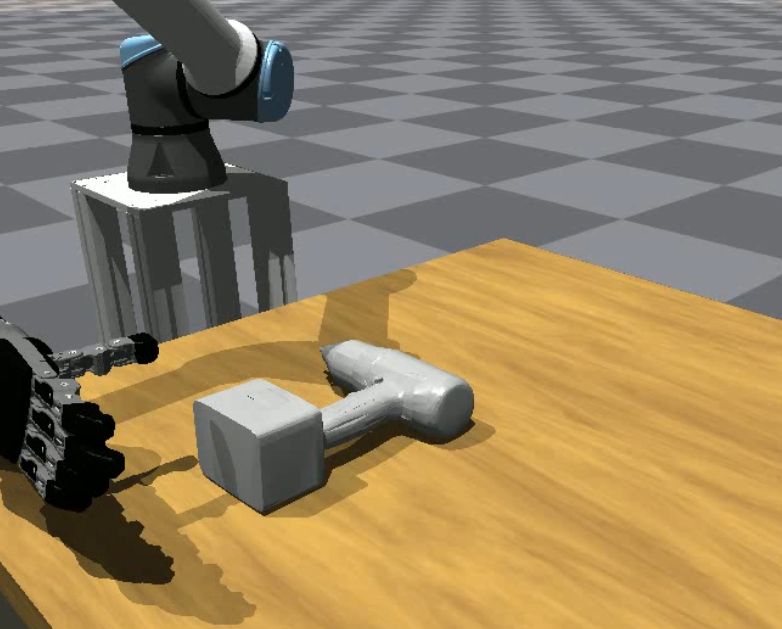}\hspace{0.5ex}
	\includegraphics[width=0.155\linewidth]{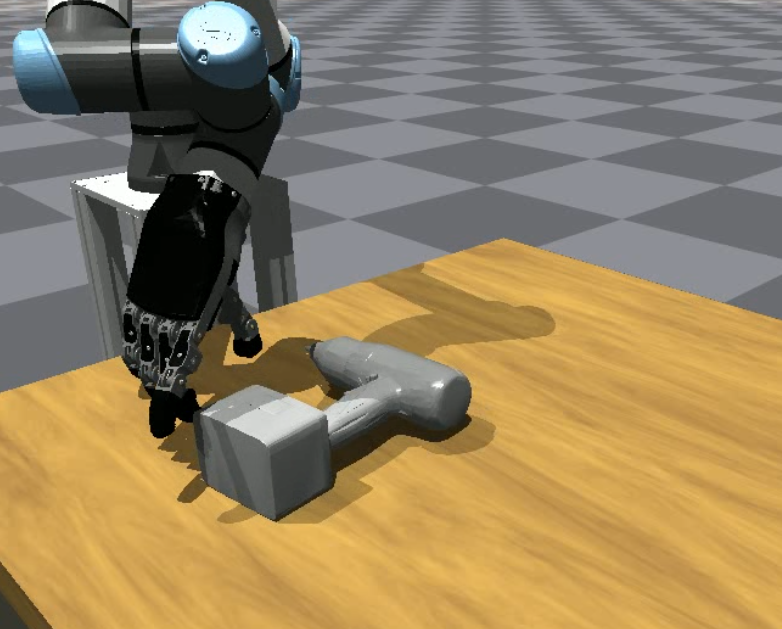}\hspace{0.5ex}
	\includegraphics[width=0.155\linewidth]{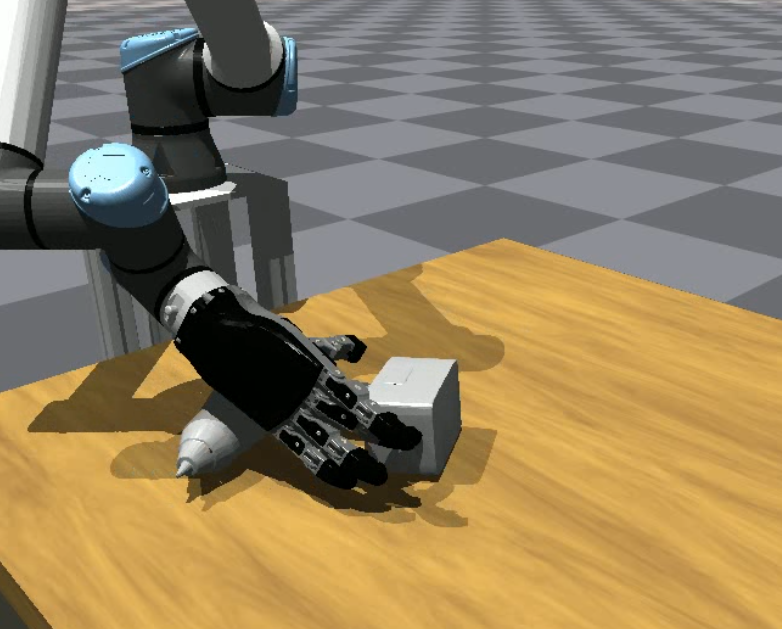}\hspace{0.5ex}
	\includegraphics[width=0.155\linewidth]{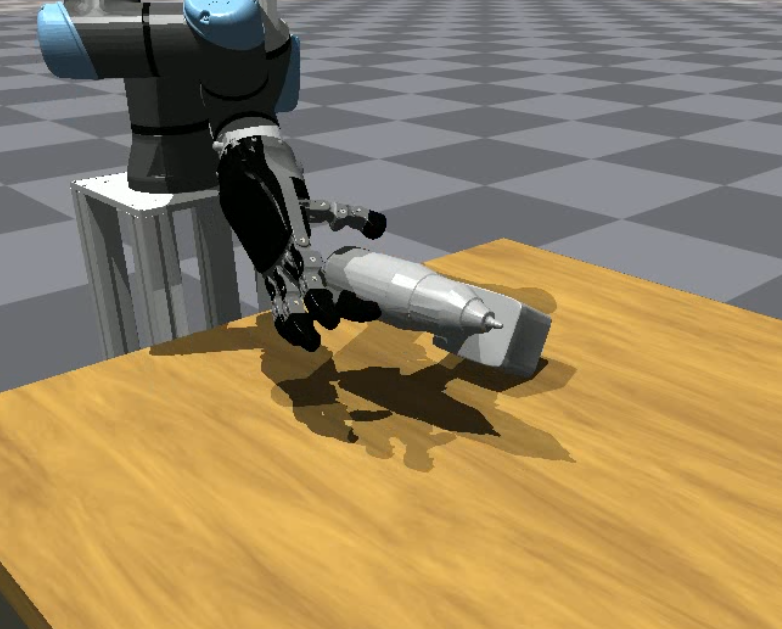}\hspace{0.5ex}
	\includegraphics[width=0.155\linewidth]{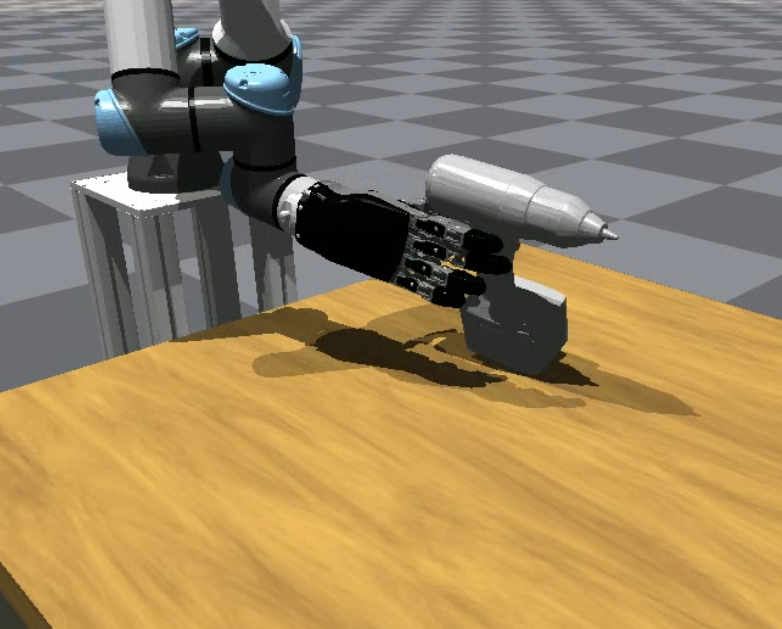}\hspace{0.5ex}
	\includegraphics[width=0.155\linewidth]{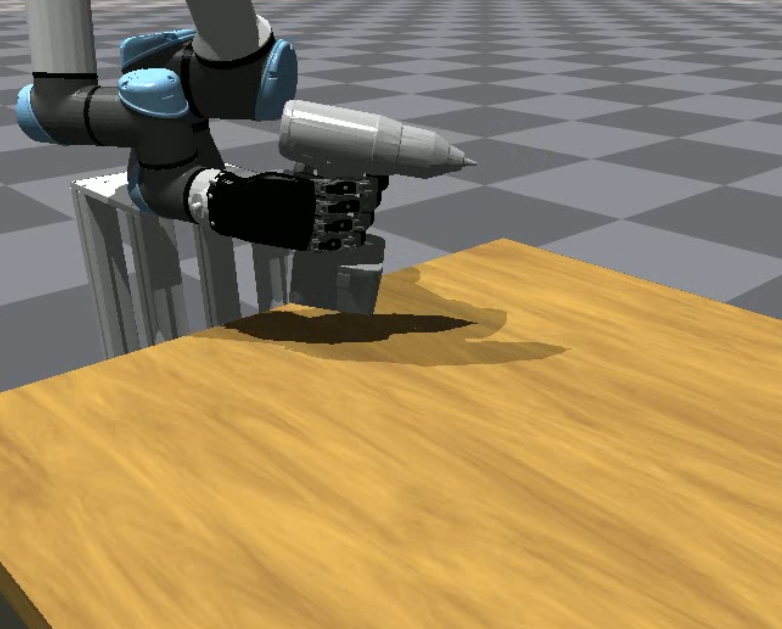}
	\\ \vspace*{1ex}
	\includegraphics[width=0.155\linewidth]{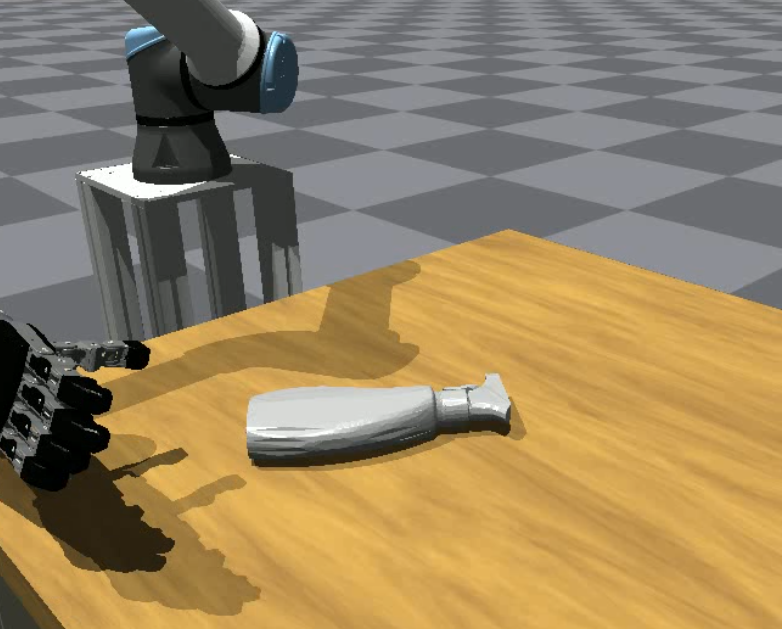}\hspace{0.5ex}
	\includegraphics[width=0.155\linewidth]{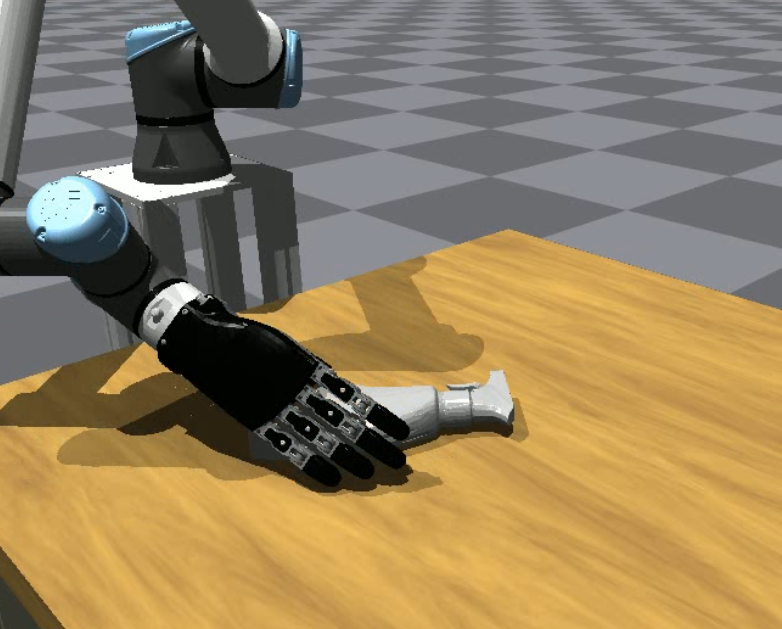}\hspace{0.5ex}
	\includegraphics[width=0.155\linewidth]{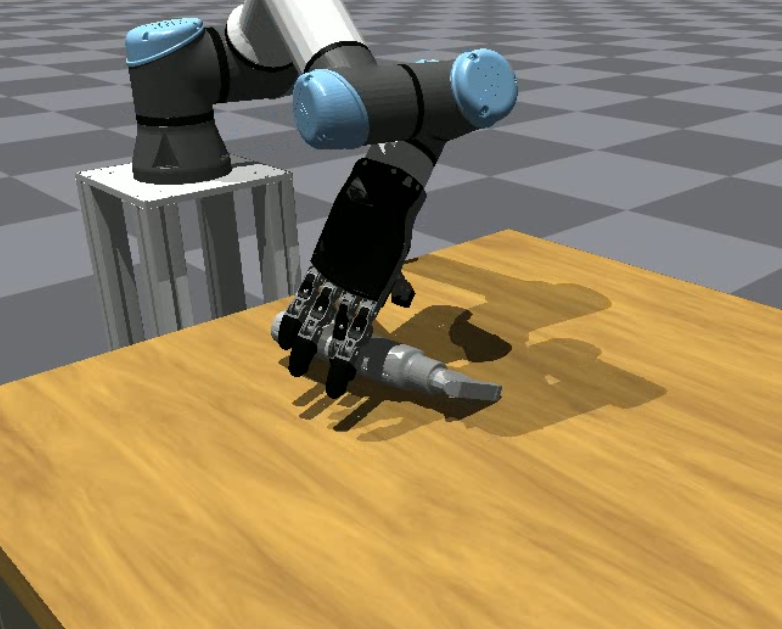}\hspace{0.5ex}
	\includegraphics[width=0.155\linewidth]{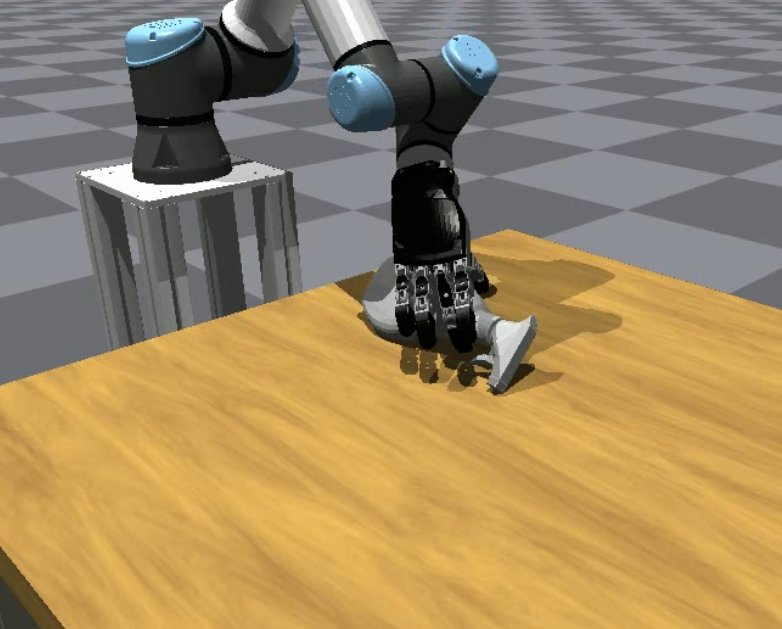}\hspace{0.5ex}
	\includegraphics[width=0.155\linewidth]{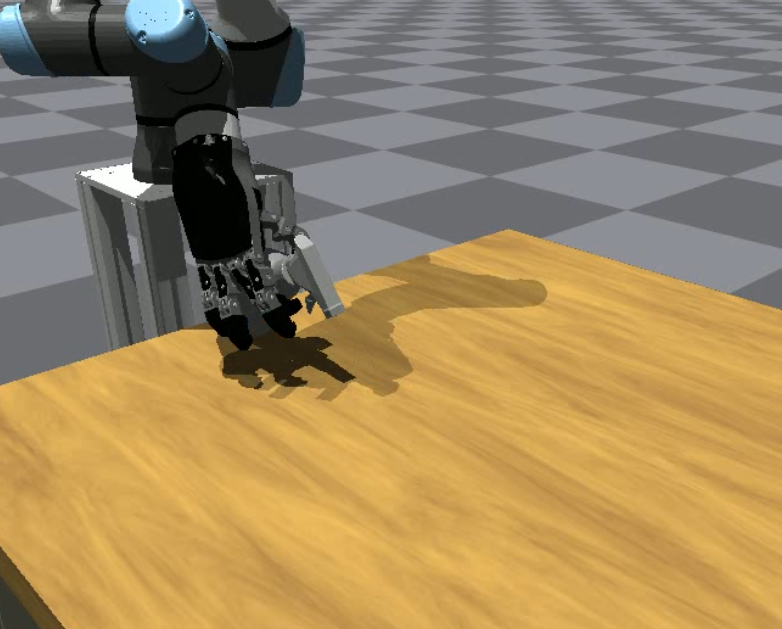}\hspace{0.5ex}
	\includegraphics[width=0.155\linewidth]{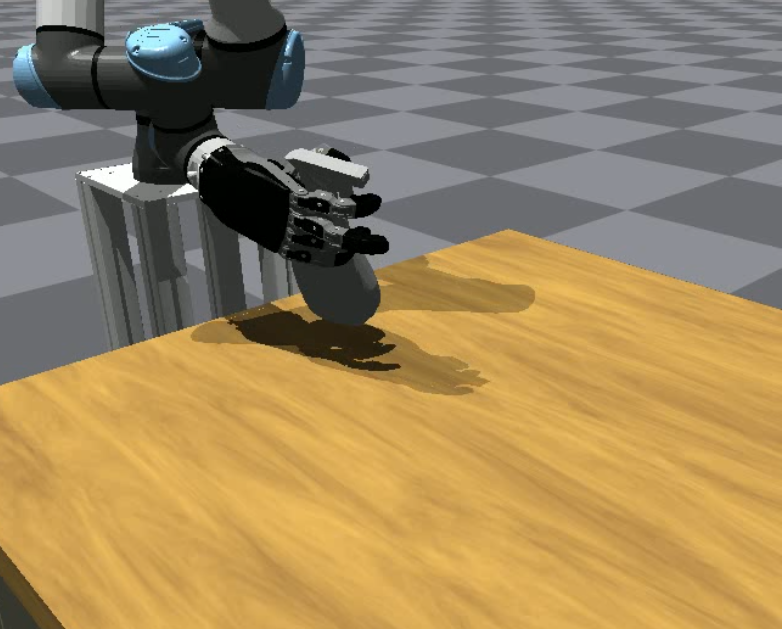}
	\\ \vspace*{1ex}
	\includegraphics[width=0.155\linewidth]{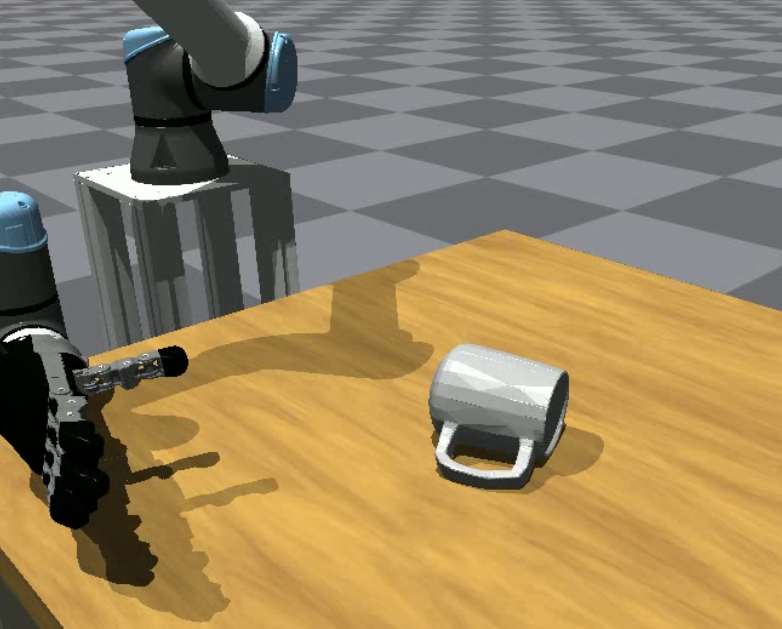}\hspace{0.5ex}
	\includegraphics[width=0.155\linewidth]{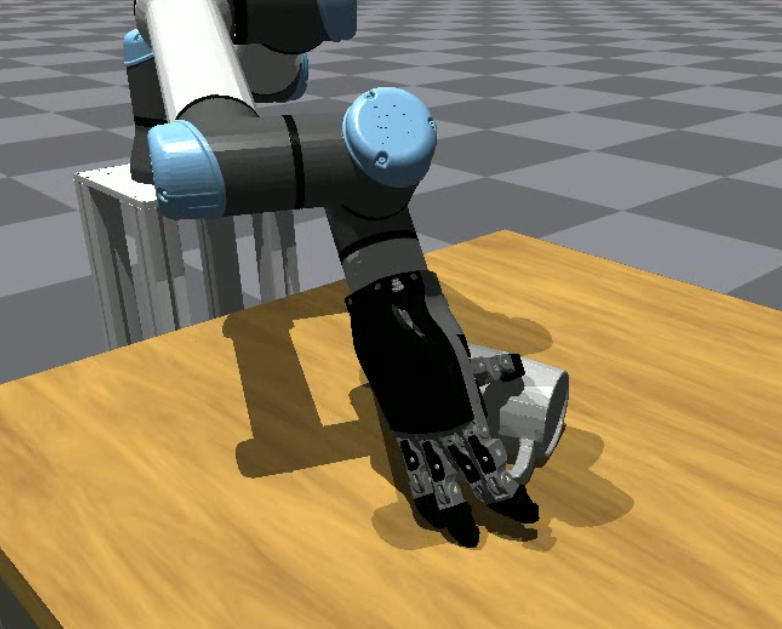}\hspace{0.5ex}
	\includegraphics[width=0.155\linewidth]{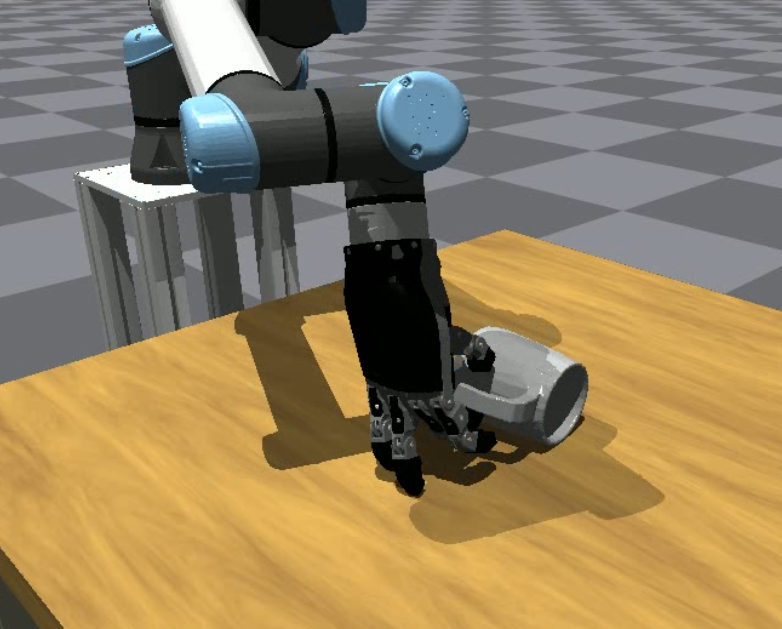}\hspace{0.5ex}
	\includegraphics[width=0.155\linewidth]{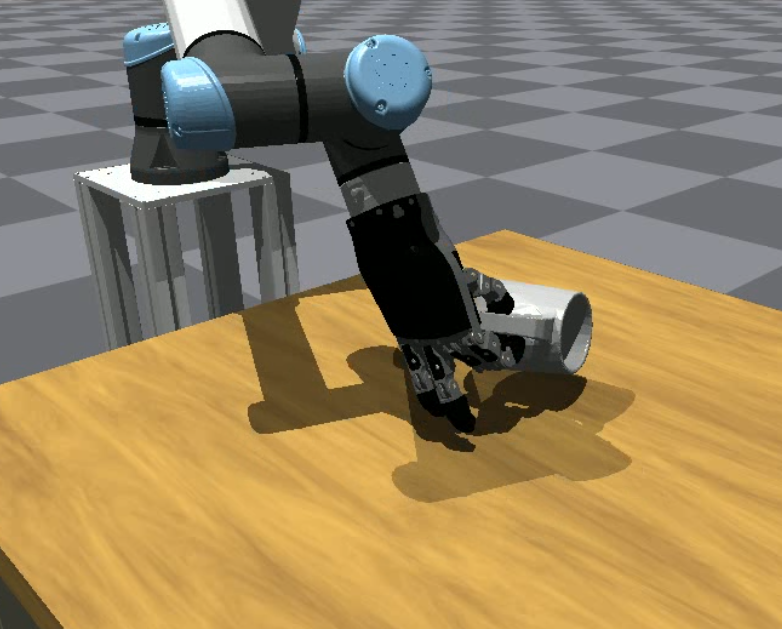}\hspace{0.5ex}
	\includegraphics[width=0.155\linewidth]{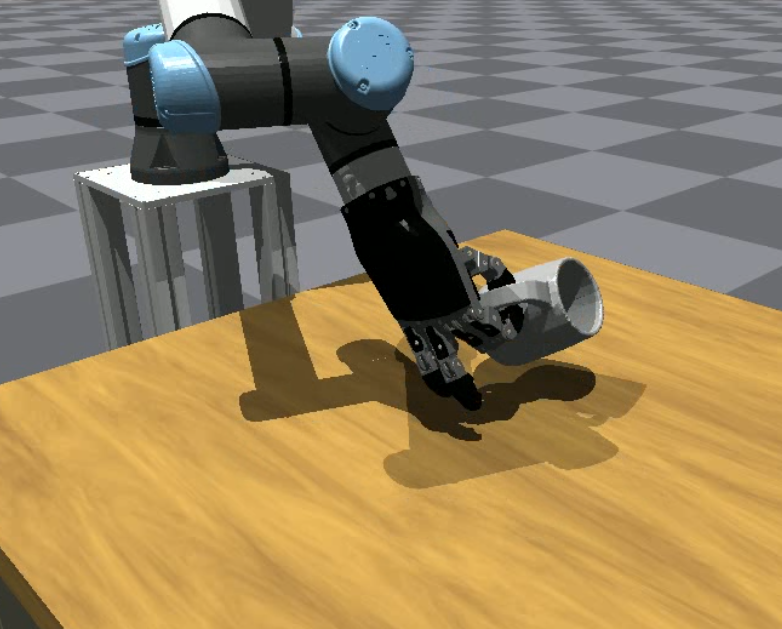}\hspace{0.5ex}
	\includegraphics[width=0.155\linewidth]{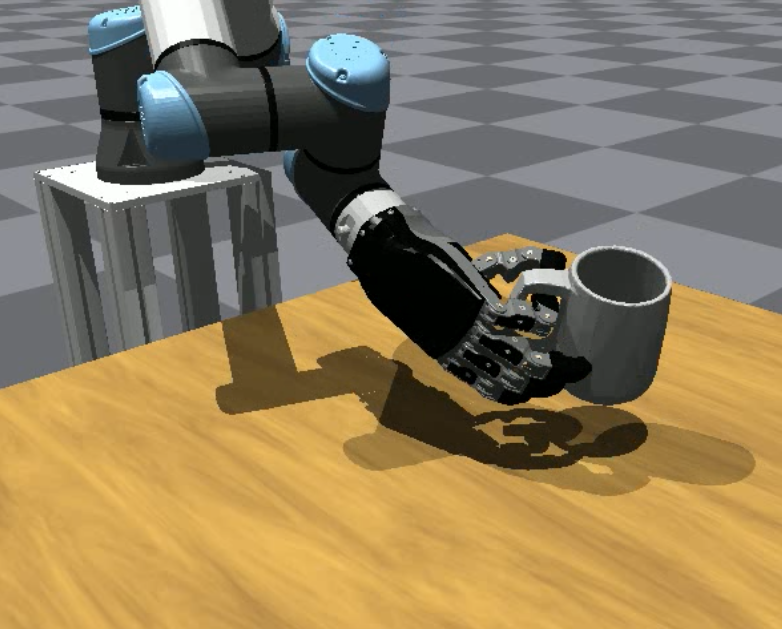}
	\caption{Constraint-based target grasp representation: Rollouts of policy manipulating unseen objects of known categories. Top to bottom: drill, spray bottle, and mug.  Note complex repositioning and reorienting behaviors for the drill and spray bottle. The mug task usually can be solved with a straightforward manipulation strategy.}
	\label{fig:rollouts_constraint}
	\vspace*{-3ex}
\end{figure*}

\begin{figure}[t]
	\centering
	\includegraphics[width=0.31\linewidth]{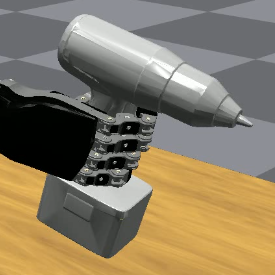}\hspace{0.5ex}
	\includegraphics[width=0.31\linewidth]{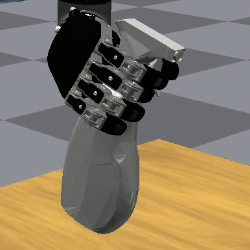}\hspace{0.5ex}
	\includegraphics[width=0.31\linewidth]{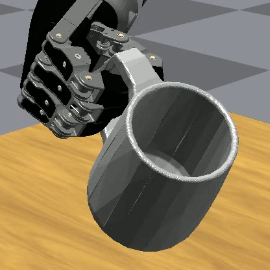}
	\\ \vspace*{1ex}
	\includegraphics[width=0.31\linewidth]{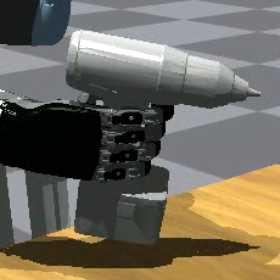}\hspace{0.5ex}
	\includegraphics[width=0.31\linewidth]{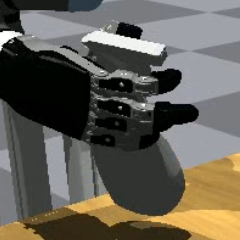}\hspace{0.5ex}
	\includegraphics[width=0.31\linewidth]{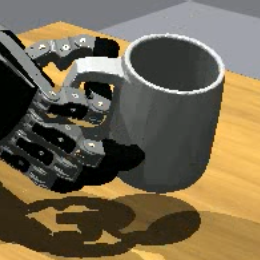}
	\caption{Close-up snapshots of functional grasps achieved by policies learned using two different grasp representations. \emph{Top:} Explicit grasp representation, carefully designed by hand. \emph{Bottom:} Constraint-based grasp representation. Given only an abstract constraint, the policy learns to produce natural-looking grasps.}
	\label{fig:grasps_constraint}
	\vspace*{-3ex}
\end{figure}

\reffig{fig:rollouts_constraint} shows example policy rollouts. The policy learns complex repositioning and reorienting behaviors to eventually satisfy the given target grasp constraints and successfully lift the objects. \reffig{fig:grasps_constraint} shows close-up snapshots of the achieved grasps using policies trained with the different grasp representations. Note that for the explicit grasp representation, the targets are carefully designed by hand. In addition, after the target pre-grasp is reached, the hand is closed and the object is lifted to confirm success. Such an approach is chosen since it is extremely challenging to supply an explicit grasp pose that has all fingers positioned perfectly and maintains consistent, tight contact with the object.

In the case of the constraint-based grasp representation, the policy has to learn grasps that enable object lifting on its own. One observation is that in the case of the constraint-based grasp representation, the spray bottles and mugs have the middle finger fully extended. For spray bottles, our intuition is that it is challenging to place the middle finger under the trigger. While trying to do so, the target index fingertip position may be disturbed, decreasing the reward. For mugs, our observation is that the policy often uses the middle finger to support the mug from the side, tilting the mug to the right, facilitating such a supporting approach. 

It is worth noting that the policy learns natural human-like ways to grasp the objects without having any explicit instructions on how to do so. We attribute this success to a generic multi-component dense reward function and a requirement to be able to lift the objects while satisfying the target grasp constraints. Thus, our methodology implicitly guides the policy towards discovering natural ways to grasp the objects, exactly in the way they are designed to be grasped by humans.

\subsection{Discussion}

The policies achieved success rates of 94\% for the explicit target grasp representation and 90\% for the constraint-based target grasp representation. In both cases, failures occurred when the object was repositioned and reoriented to a workspace location with low manipulability, leading the policy to be stuck in repetitive behavior. To address this issue, an additional later stage in the training curriculum could be introduced, where most of the reward components have a negative range. Having all reward components with a negative range from the start resulted in very slow learning.

While the proposed method consistently learned policies with high success rates in simulation, transferring the approach to the real world is not straightforward. The main limitation of the proposed approach is its reliance on frequent and accurate estimation of the target object pose. In the real world, in the presence of the robotic hand, 6D object pose estimation is challenging~\cite{AminiPB:IAS22}. Although we introduce some degree of noise to all measurements, that does not fully model measurements of significantly lower accuracy in the presence of occlusions. In addition, we utilize certain privileged information, such as distance from fingers to the object surface. Although object shape can be reconstructed~\cite{Rodriguez_2020}, it still would be subject to major inaccuracies.

Considering these limitations, we formulate a three-step strategy that could be implemented to transfer the presented approach to the real-world scenarios:
\begin{enumerate}
	\item Transfer the learned behaviors to a model that does not require distances from fingers to the object surface. Policy distillation~\cite{Wojciech_2019} or newer methods~\cite{Mosbach_2024} can be applied to achieve that.
	\item Introduce a reward term that penalizes object occlusion. It was shown to be effective when applied to arbitrary grasping~\cite{Mosbach_2024}. Additionally, distort the object-related observations proportionally to the occlusion. 
	\item Apply the approach to the real system. Before manipulation begins, object shape can be reconstructed~\cite{Rodriguez_2020, Han_2021, Yang_2021, Zhou_2024}, enabling defining the target functional grasp~\cite{Rodriguez_2018, Wu_2023, Zhang_2023, Wei_2024}. Finally, the 6D pose of the object can be continuously estimated~\cite{Deng_2020, Hu_2020, Hofer_2021, AminiPB:IAS22, Wang_2024}. Data-efficient real robot DRL~\cite{Pavlichenko_2022} can be applied to update the weights of the last layer of the model to compensate for differences between simulation and the real world.
\end{enumerate}

The presented three-step approach gradually addresses the challenges of transferring the method to the real world. That includes eliminating the need for privileged observations and minimizing object occlusions, before finally applying the learned policy to the real robot. We believe that it is a promising work direction that can be realized with recent developments in areas of perception and manipulation learning.
\section{Conclusion}
\label{sec:Conclusion}

In this article, we presented a DRL approach for dexterous categorical pre-grasp manipulation for functional grasping with an anthropomorphic hand. We introduced a dense multi-component reward function and a curriculum to quickly learn a single policy for dexterous manipulation of complex objects of three categories. We proposed two target grasp representations: explicit and a more abstract, constraint-based one.

Our experiments demonstrated that learning with our approach reliably converges and produces policies that achieve high success rates, even for previously unseen object instances of known categories. Complex pre-grasping strategies such as repositioning, reorienting, regrasping, and up-righting the object have been learned.

Ablation studies confirmed the importance of the proposed multi-component reward function and the curriculum. Our method utilizes a high-performance GPU-based simulation, and the policies for both target grasp representations were learned on a single GPU in less than three hours. The policy using an explicit target grasp representation achieved a 94\% success rate for functional grasping of novel object instances. The policy utilizing a constraint-based target grasp representation achieved a 90\% test success rate while simultaneously learning human-like grasp configurations solely from the provided functional grasp constraints. Stable convergence of both policies and resulting high success rates demonstrate the generality of the proposed learning pipeline.

\section*{Acknowledgments}
We would like to thank Malte Mosbach for providing the source code of his work on object geometry representations for grasping~\cite{Mosbach_2022} as a base for this research.

\bibliographystyle{IEEEtran}
\bibliography{bibliography}


\vspace{11pt}

\vspace{-33pt}
\begin{IEEEbiography}[{\includegraphics[width=1in,height=1.25in,clip,keepaspectratio]{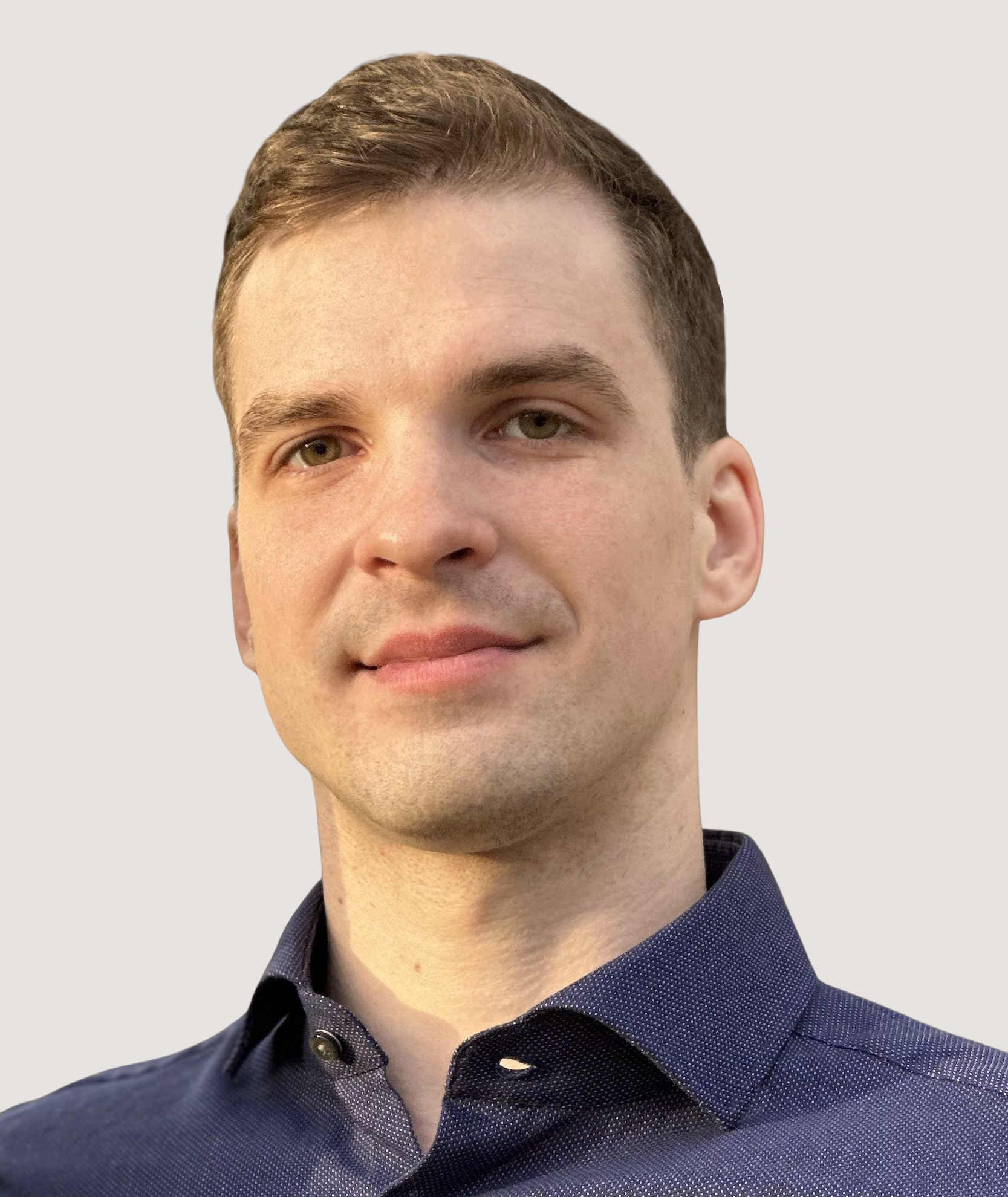}}]{Dmytro Pavlichenko}
received his M.S. degree in Computer Science
from the University of Bonn, Germany in 2017. Since 2017, he has been conducting his doctoral studies at the Autonomous Intelligent Systems group at the University of Bonn, Germany. During the doctoral studies, he has actively participated in international projects such as CENTAURO and Learn2Grasp, and contributed to winning several robotics competitions including the MBZIRC 2017 and RoboCup 2016, 2017, 2018, 2019, 2022, 2023. Dmytro Pavlichenko is the author of numerous scientific publications in the field of robotics. His research interests include machine learning, robotic arm trajectory tracking, and manipulation planning.
\end{IEEEbiography}

\vspace{11pt}
\vspace{-33pt}

\begin{IEEEbiography}[{\includegraphics[width=1in,height=1.25in,clip,keepaspectratio]{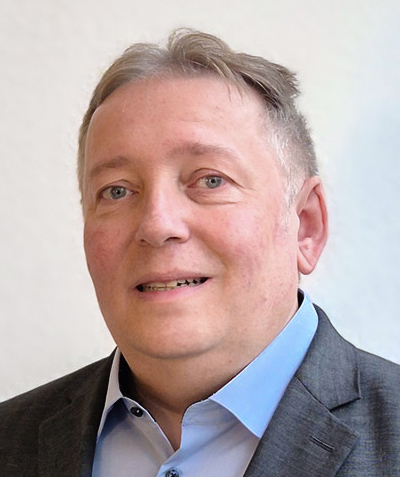}}]{Sven Behnke}
is since 2008 professor for Autonomous Intelligent Systems at the 
University of Bonn and director of the Computer Science Institute VI -- Intelligent Systems and Robotics. 
He heads the research area Embodied AI of the Lamarr Institute for Machine Learning and Artificial Intelligence and is ELLIS Fellow.
Prof. Behnke received his M.S. degree in Computer Science (Dipl.-Inform.) in 1997 from Martin-Luther-Universit\"at Halle-Wittenberg. 
In 2002, he obtained a Ph.D. in Computer Science (Dr. rer. nat.) from Freie Universit\"at Berlin. He spent the year 2003 as postdoctoral 
researcher at the International Computer Science Institute, Berkeley, CA. From 2004 to 2008, Prof. Behnke headed the Humanoid 
Robots Group at Albert-Ludwigs-Universit\"at Freiburg. His research interests include cognitive robotics, computer vision, and machine learning. 
\end{IEEEbiography}

\vfill

\end{document}